%% file: main.tex
\documentclass[10pt,twocolumn,letterpaper]{article}

\usepackage{cvpr}              
\usepackage{multirow}
\usepackage{booktabs}
\usepackage{tabularx}
\newcolumntype{Y}{>{\centering\arraybackslash}X}
\usepackage{algorithm}
\usepackage{algpseudocode}

\usepackage{amssymb}  
\usepackage[utf8]{inputenc}
\usepackage[T1]{fontenc}
\usepackage{tabularx}
\usepackage{booktabs}
\usepackage{comment}
\input{preamble}
\definecolor{cvprblue}{rgb}{0.21,0.49,0.74}
\usepackage[pagebackref,breaklinks,colorlinks,allcolors=cvprblue]{hyperref}

\usepackage[table]{xcolor} 
\definecolor{babyblue}{RGB}{173, 216, 230} 

\definecolor{CBmagenta}{RGB}{210, 50, 150}

\def\confName{CVPR}
\def\confYear{2026}

\title{Lifelong Imitation Learning with \\ Multimodal Latent Replay and Incremental Adjustment}

\author{
Fanqi Yu$^{1,3}$ \quad
Matteo Tiezzi$^{4\dagger}$ \quad
Tommaso Apicella$^{4\dagger}$ \quad
Cigdem Beyan$^{1,2}$ \quad
Vittorio Murino$^{1,2}$ \\
$^{1}$AI for Good (AIGO), Istituto Italiano di Tecnologia, Genoa, Italy \\
$^{2}$Department of Computer Science, University of Verona, Verona, Italy \\
$^{3}$DITEN, University of Genoa, Genoa, Italy \\
$^{4}$PAVIS, Istituto Italiano di Tecnologia, Genoa, Italy \\
$^{\dagger}$These authors contributed equally as second authors.
}

\begin{document}
\maketitle
\input{sec/0_abstract}    
\input{sec/1_intro}

\input{sec/2_related_work}
\input{sec/3_preliminaries}
\input{sec/3_proposed_method}
\input{sec/4_experiments}
\input{sec/5_conclusion}

{
    \small
    \bibliographystyle{ieeenat_fullname}
    \bibliography{main}
}

\input{sec/X_suppl}

\end{document}

%% file: sec/0_abstract.tex
\begin{abstract}
We introduce a lifelong imitation learning framework that enables continual policy refinement across sequential tasks under realistic memory and data constraints. Our approach departs from conventional experience replay by operating entirely in a multimodal latent space, where compact representations of visual, linguistic, and robot's state information are stored and reused to support future learning. To further stabilize adaptation, we introduce an incremental feature adjustment mechanism that regularizes the evolution of task embeddings through an angular margin constraint, preserving inter-task distinctiveness.  
Our method establishes a new state of the art in the LIBERO benchmarks, achieving 10–17 point gains in AUC and up to 65\% less forgetting compared to previous leading methods. Ablation studies confirm the effectiveness of each component, showing consistent gains over alternative strategies. The code is available at: \url{https://github.com/yfqi/lifelong_mlr_ifa}.

\end{abstract}

%% file: sec/1_intro.tex
\section{Introduction}
\label{sec:intro}
Imitation Learning (IL) enables agents, such as robots, to learn behaviors by observing and mimicking human demonstrations~\cite{li2025robotic,stepputtis2020language,jang2022bc,xie2024decomposing,hussein2017imitation}. However, real-world environments are dynamic, with new objects, goals, and contexts constantly emerging. For example, a household robot may encounter unfamiliar kitchen tools, rearranged furniture, or previously unseen tasks. To operate effectively, agents must continuously learn and adapt throughout their lifetime. Standard IL typically assumes a fixed set of tasks and does not account for new tasks or variations in the environment~\cite{zare2024survey,zheng2022imitation}. Lifelong Imitation Learning (LIL) addresses this limitation by enabling agents to continuously acquire new skills while retaining previously learned behaviors, even as the set of tasks evolves~\cite{zheng2025towards,tsuji2025survey,liu2023libero,yao2025think}. LIL focuses on preventing catastrophic forgetting~\cite{kemker2018measuring}, allowing knowledge from earlier tasks to be reused for new ones, and supports learning from an unbounded or dynamically growing sequence of tasks, rather than a fixed predefined set~\cite{wan2024lotus,chaudhry2019tiny,gao2021cril,roy2025m2distill}.
\begin{figure}[t!]
    \centering
    \includegraphics[width=1.\linewidth,trim=1.4cm 0 1.4cm 0cm, clip]{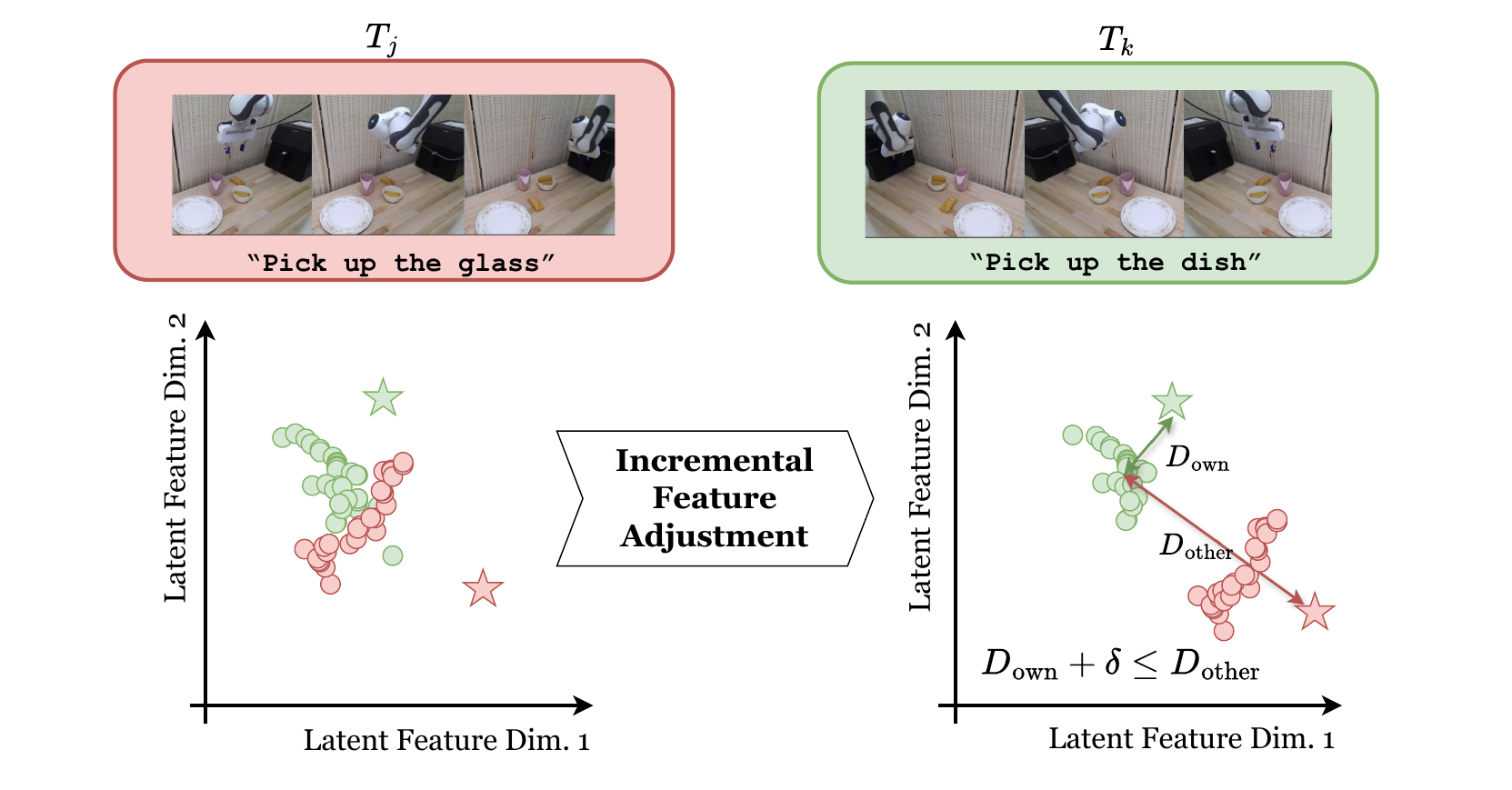}
    \caption{Illustration of Incremental Feature Adjustment (IFA). The figure displays a 2D projection of the global latent representations $\mathbf{g}$ during policy rollout for two related tasks, $T_j$ (previously learned) and $T_k$ (newly learned). The stars ($\mathbf{\color{red}\star}$ for $T_j$ and $\mathbf{\color{green}\star}$ for $T_k$) represent the stable language reference embeddings $\mathbf{h}^{(r)}$, while the circles ($\mathbf{\color{red}\bullet}$ for $T_j$ and $\mathbf{\color{green}\bullet}$ for $T_k$) are the global embeddings $\mathbf{g}$. (Left) Without IFA, the new task's embeddings ($\mathbf{\color{green}\bullet}$) exhibit representation drift by clustering close to the old task's embeddings ($\mathbf{\color{red}\bullet}$). (Right) With IFA, the loss $\mathcal{L}_{\text{IFA}}$ enforces a constraint on the distances: the distance to the own reference ($D_{\text{own}}$) plus a margin $\delta$ must be less than or equal to the distance to the other task's reference ($D_{\text{other}}$). This mechanism forces the $\mathbf{g} (T_k)$ ($\mathbf{\color{green}\bullet}$) cluster away from $\mathbf{h}^{(r)}(T_j)$ ($\mathbf{\color{red}\star}$)  and closer to $\mathbf{h}^{(r)}(T_k)$ ($\mathbf{\color{green}\star}$) , achieving inter-task disentanglement.}
    \vspace{-5mm}
    \label{fig:teaser}
\end{figure}

However, many practical methods rely on additional assumptions to prevent forgetting. For instance, some approaches assume that the task identifier is available during both training and evaluation, allowing task-specific networks or adapters to isolate knowledge for each task \cite{liu2023tail}. In this way, these methods retain performance on previously learned tasks while sequentially acquiring new ones. In contrast, fully task-agnostic LIL methods do not require task identifiers at test-time (referred to as \textit{task-ID agnostic} throughout this paper) and typically rely on strategies such as experience replay (also called rehearsal-based) \cite{liu2023libero,wan2024lotus,gao2021cril}, or multi-task distillation \cite{roy2025m2distill}.

Recent LIL methods have increasingly leveraged large pretrained models, such as Vision-Language Models (e.g., CLIP~\cite{radford2021learning}) and Large Language Models (e.g., GPT~\cite{achiam2023gpt}). However, directly using these models in their pretrained form on natural images or text is often insufficient for complex imitation learning tasks. To address this, several approaches, e.g.,~\cite{wan2024lotus,lee2024incremental,liu2023tail} adopt a two-stage paradigm: a pretraining stage (also called the base task stage) followed by a lifelong learning stage. This strategy leverages the rich representations learned during pretraining while enabling continuous adaptation to new tasks. 
During the lifelong learning stage, some prior methods still employ parameter-efficient fine-tuning (PEFT) (e.g.,~\cite{ding2023parameter,xin2024parameter,lialin2023scaling}) to adapt encoders to novel tasks~\cite{liu2023tail,lee2024incremental}. 

In this study, we introduce a rehearsal-based method that applies a Multimodal Latent Replay (MLR), which stores joint compact latent representations that encapsulate vision, language, state (e.g., robot's orientation, position) modalities with control commands. When a new task is encountered, its latent representations may overlap those of previously learned tasks~\cite{huang2024class}, leading to interference in the shared embedding space. To address this, we propose Incremental Feature Adjustment (IFA), a representation-level regularization strategy that maintains reference embeddings for past tasks, such as language-based task embeddings or centroids of aggregated feature representations (See Fig.~\ref{fig:teaser}). IFA is a loss on angular distances to encourage the current task’s representation to remain distinct from previous tasks while maintaining alignment with its own reference. The loss margin is adaptively scaled based on inter-task similarity, reducing sensitivity to dataset-specific feature variations and eliminating the need for manual tuning. By progressively repelling the current task from old-task anchors, IFA preserves inter-task separability and mitigates interference during lifelong learning.

Our approach uses a frozen CLIP encoder \cite{radford2021learning}, demonstrating that effective LIL can be achieved without fine-tuning the backbone during the lifelong learning stage, unlike \cite{liu2023tail,lee2024incremental}, which rely on PEFT (e.g., vision adapters). Different from rehearsal-based LIL methods, e.g., \cite{liu2023libero,wan2024lotus}, which store raw data (e.g., scenes and trajectories), we employ compact MLR, which is memory-efficient. By combining MLR with IFA, our method achieves robust lifelong learning in a task-ID agnostic manner, offering a simpler alternative to distillation \cite{roy2025m2distill} or generative approaches \cite{gao2021cril}.

Experimental analysis on three different datasets demonstrates that the proposed MLR and IFA modules provide noticeable improvements over existing state-of-the-art (SOTA). Ablation studies show that each component contributes meaningfully, outperforming alternative designs or variations built on different assumptions.

The contributions of this work can be summarized as:
\begin{itemize}
    \item A multimodal latent replay framework for learning new tasks using a model with pre-trained unimodal encoders (visual, language, and state). By storing latent features instead of raw data, our method reduces the memory footprint of previously learned tasks and mitigates forgetting.
    \item An Incremental Feature Adjustment module that separates the latent representations of old and new tasks. Specifically, we modulate the margin between representations based on the angular distance between old and new task embeddings, adapting the strength of the loss according to the semantic similarity between tasks.
\end{itemize}

%% file: sec/2_related_work.tex
\section{Related Work}
\label{sec:relatedWork}

Lifelong Imitation Learning (LIL) aims to enable robotic agents to acquire new skills over time without forgetting previously learned ones, a challenge commonly referred to as catastrophic forgetting. To address this problem, some studies have employed rehearsal-based techniques (also known as experience replay)~\cite{wan2024lotus,liu2023libero,gao2021cril}, which replay stored past data to help the model maintain performance on earlier tasks.
For instance, \cite{liu2023libero}, building upon ER~\cite{chaudhry2019tiny}, maintains a selection of past trajectories and interleaves them with new ones from the current task during training.
In comparison, CRIL~\cite{gao2021cril} employs generative adversarial networks (GANs) to synthesize the first frame of each trajectory and uses an action-conditioned video prediction model to generate subsequent frames based on the current states and actions. LOTUS~\cite{hu2022lora} allows robots to incrementally acquire new skills by identifying recurring patterns within unsegmented demonstrations. These skills are extracted via an open-vocabulary vision model and orchestrated by a meta-controller, enabling the agent to solve complex, long-horizon manipulation tasks. Despite their effectiveness, rehearsal-based approaches are often sensitive to the replay ratio and the similarity between new and previously learned tasks, resulting in variable levels of forgetting.

Another line of work focuses on progressive model expansion, where additional parameters or modules are introduced as new tasks arrive, enabling the architecture to grow adaptively and incorporate new knowledge. BUDS~\cite{zhu2022bottom} proposes a method for discovering reusable skills in robotic manipulation without relying on pre-segmented demonstrations. Using a bottom-up strategy, it autonomously identifies and organizes skills from long-horizon, unsegmented demonstrations, allowing robots to handle complex manipulation tasks effectively.
In contrast, TAIL~\cite{liu2023tail} applies several Parameter-Efficient Fine-Tuning (PEFT) methods, such as LoRA, to CLIP encoders, creating a specific adapter for each new task. A key limitation of this approach is that it requires the task identifier at test time, which may not always be available in practical scenarios.
Another adapter-based method, ISCIL~\cite{lee2024incremental}, addresses this limitation by promoting knowledge sharing through the incremental learning of reusable skills across multiple demonstrations. These skills are stored in a prototype-based memory, enabling sample-efficient task adaptation and better generalization in non-stationary LIL environments.
Overall, progressive approaches generally rely on explicit stage or task identification during evaluation and often face challenges in generalizing to unseen tasks.

Differently, M2Distill~\cite{roy2025m2distill} is a distillation-based LIL method that preserves a consistent latent space across vision, language, and action modalities as new tasks are learned. It regulates distribution shifts between consecutive learning steps via multimodal distillation and Gaussian Mixture Model (GMM) policy alignment, ensuring that previously acquired skills are retained while new ones are integrated efficiently.

Our approach can be categorized as rehearsal-based, but it differs from the above-mentioned methods by replaying multimodal latent representations, which are the compositions of vision, language, the state of the robot, and action modalities. Considering that high similarity between new and previously learned tasks can degrade the performance of standard rehearsal-based methods, we introduce a incremental feature adjustment mechanism based on angular distance. Compared to other approaches, our pipeline is much simpler: it does not rely on knowledge distillation, does not use any PEFT methods, and instead operates on completely frozen backbones. Moreover, unlike methods such as LOTUS~\cite{wan2024lotus}, our approach does not require an open-vocabulary vision encoder to achieve superior performance.

%% file: sec/3_preliminaries.tex
\section{Methodology}

\begin{figure*}[t!]
    \centering
    \includegraphics[width=\linewidth]{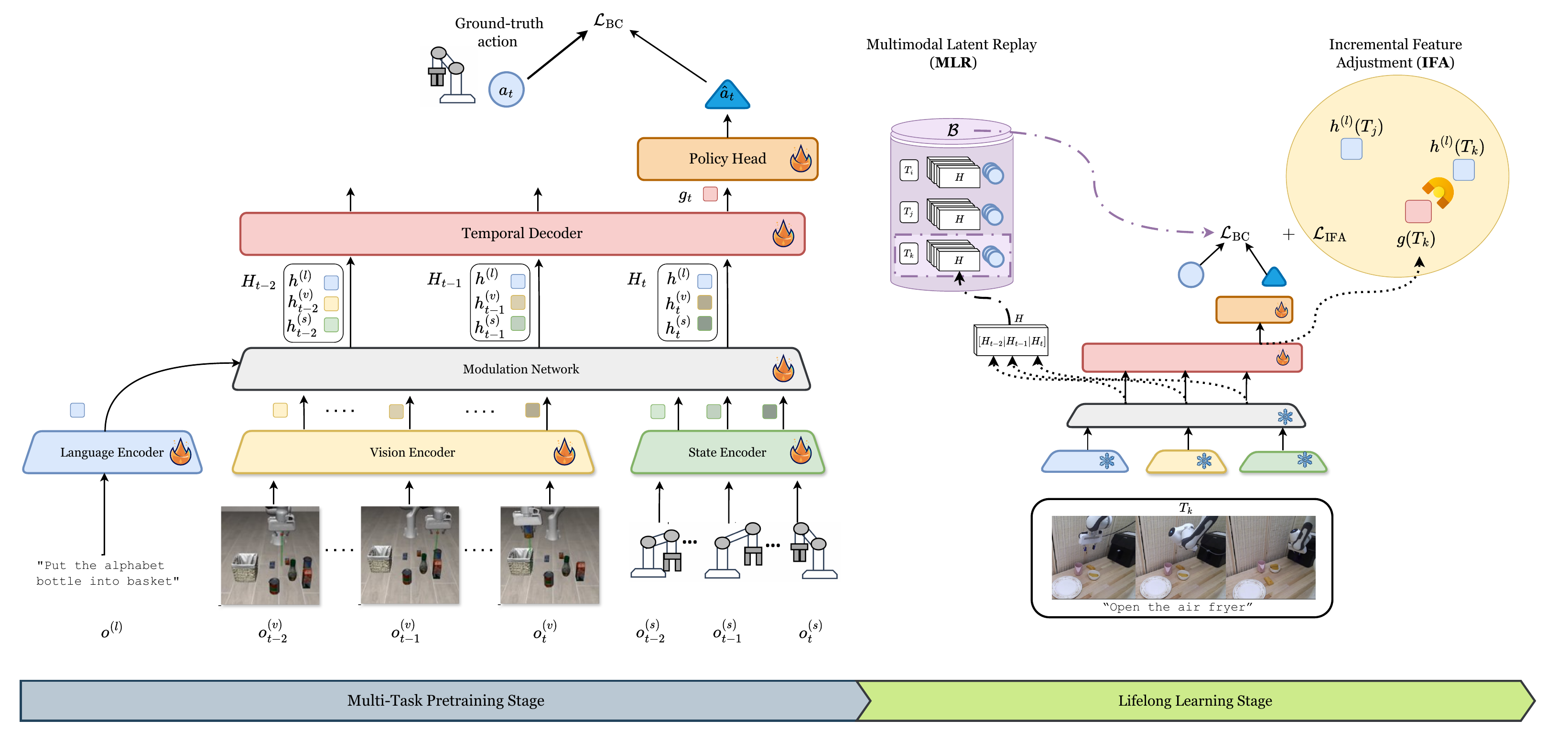}
    \caption{Our method is a general multimodal architecture composed of modality-specific encoders (language, vision, and state), a modulation network, a temporal decoder, and a policy head. 
    During the pretraining phase, all architecture modules are trained. In the lifelong learning phase, only the temporal decoder and policy head are updated using both the new task data and the samples stochastically stored in the replay buffer. The buffer stores the multimodal features, output of the modulation layer. The model is jointly supervised by both the Behavior Cloning and the Incremental Feature Adjustment loss, processing the current task and previously stored tasks.} 
    \label{fig:architecture}
    \vspace{-0.5em}
\end{figure*}

\subsection{Problem Formulation}

We frame the LIL problem as consisting of two sequential phases: a \emph{multi-task pre-training stage} and a \emph{lifelong learning stage}, following prior work \cite{wan2024lotus,lee2024incremental,liu2023tail}. In this subsection, we provide an overview of this problem formulation, while the next subsection details our approach.

\textbf{Multi-Task Pre-Training.}
Let $\mathcal{T}' = \{ T'_j \}_{j=1}^{J}$ denote a set of pre-training tasks. 
Each task $T'_j$ is associated with a set of $N_j$ expert demonstrations (e.g., trajectories collected from human operators or expert policies that successfully perform the task) $\mathcal{D}_j = \{ \tau_{j,i} \}_{i=1}^{N_j}$, where each trajectory $\tau_{j,i} = \{ (o_t, a_t) \}_{t=1}^{L_j}$ 
consists of a sequence of $L_j$ timesteps of observation-action pairs collected from expert rollouts.
Here, the observation $o_t$ typically includes sensory observations such as different views of images, proprioceptive signals and language description of the task to be performed, while $a_t$ represents the expert’s control command (ground truth action). 

The policy $\pi_\theta(\hat{a}_t \mid o_{\le t})$, parameterized by $\theta$, 
predicts the action $\hat{a}_t$ conditioned on the most recent $L_j$ observations, denoted  with $o_{\le t} \triangleq (o_{t-L_j+1}, \cdots, o_t)$,
and is trained to imitate expert actions using a behavioral cloning (BC) objective~\cite{pomerleau1988alvinn}:
\begin{equation}
    \min_{\theta} 
    \sum_{j=1}^{J} 
    \mathbb{E}_{(o_{\le t}, a_t) \sim \mathcal{D}_j}
    \left[
        \mathcal{L}_{\text{BC}}
        \big(
            \pi_\theta(\hat{a}_t \mid o_{\le t}),\, a_t
        \big)
    \right],
\end{equation}
where $\mathcal{L}_{\text{BC}}$ denotes a supervised imitation loss 
(e.g., negative log-likelihood or mean-squared error).
This pre-training phase is performed jointly over all tasks in an \emph{offline} multi-task setting, 
serving to establish shared representations and behavioral priors before lifelong learning begins. 

\textbf{Lifelong Learning.}
After the pre-training stage, the agent encounters a sequence of unseen tasks  $\mathcal{T} =\{ T_{1}, T_{2}, \ldots, T_{K} \}$, 
where each task $T_{k}$ arrives sequentially, with $\mathcal{T} \cap \mathcal{T}' = \emptyset$. At task ${k}$, the agent has access to a set of expert demonstrations 
$\mathcal{D}_k = \{ \tau_{k,i} \}_{i=1}^{N_k}$ from the current task, 
and a limited replay buffer $\mathcal{B}$. 
The goal is to adapt the policy $\pi_\theta$ on the union of current and replayed samples 
while avoiding catastrophic forgetting.

Formally, the learning objective at task $k$ is defined as:
\begin{equation}
    \min_{\theta}
    \mathbb{E}_{(o_{\le t}, a_t) \sim (\mathcal{D}_k \cup \mathcal{B})}
    \left[
        \mathcal{L}_{\text{BC}}
        \big(
            \pi_\theta(\hat{a}_t \mid o_{\le t}),\, a_t
        \big)
    \right].
\end{equation}

%% file: sec/3_proposed_method.tex
\subsection{Our Approach}
\label{sec:ourapproach}
We propose a LIL framework that introduces two complementary components for the lifelong learning stage: (1) Multimodal Latent Replay (MLR), which replays compact multimodal latent representations instead of raw trajectories, and (2) Incremental Feature Adjustment (IFA), which regularizes cross-task representations to maintain stability during continual adaptation. Below, we will first briefly describe the overall information flow in the neural architecture, including our pretraining, and then describe the main components we propose for the lifelong learning stage.

\textbf{Base Policy Architecture.}
The policy $\pi_\theta$ processes multimodal observations {$o_t = \{o_t^{(v)},  o^{(l)}, o_t^{(s)}\}$},
corresponding to the modalities of visual, language , and state. 
Each modality is encoded into a latent feature $\mathbf{h}_t^{(m)} \in \mathbb{R}^E$, with $m \in \{v,l,s\}$ and where $E$ denotes the embedding size, via modality-specific encoders,
modulated by a proper network and concatenated  over the past $L$ time steps to form a multimodal sequence $\mathbf{H} \in \mathbb{R}^{M \times L \times E}$, where $M$ denotes number of modalities and  $L$ the temporal length of the rollout, respectively. A {temporal decoder} 
maps $\mathbf{H}$ into a \textit{global latent representation} $g_t \in \mathbb{R}^E$, which is then decoded by the {policy head} to predict the next action $\hat{a}_t$. During pre-training, all modules are trainable. 
Fig. \ref{fig:architecture} illustrates the proposed architecture together with the training pipeline. The corresponding implementation details are described in the following text and in Sec. \ref{sec:impDet}.

\textbf{Multimodal Latent Replay (MLR).}
During the lifelong learning phase, only the temporal decoder and policy head are trainable; all other networks remain frozen as learned during pre-training.
We leverage a replay-based method, MLR, that instead of storing past trajectories with the associated ground truth action $(o_{\le t}, a_t)$ as in experience replay~\cite{liu2023libero}, maintains a compact buffer $\mathcal{B}$ of multimodal latent representations:

\begin{equation}
\mathcal{B} = \{ (\mathbf{H}_n, a_n) \}_{n=1}^{N_B},
\label{eq:buffer_definition}
\end{equation}
where $\mathbf{H}_n$ is the concatenated latent produced by the frozen encoders, and $N_B$ denotes the total buffer capacity.
When learning a new task, replayed latents from $\mathcal{B}$ are combined with the current data $\mathcal{D}_k$ to form the imitation objective. 
MLR effectively stabilizes training and mitigates catastrophic forgetting
without accessing raw sensory inputs (e.g., high-dimensional images that occupy large storage space, whereas latent trajectories are much smaller). 

\textbf{Incremental Feature Adjustment.}
To counter representation drift across tasks, 
we introduce \emph{Incremental Feature Adjustment (IFA)},
which regularizes the relationship between latent representations of old and new tasks.
Let $g_t(T_i)$ denote the global latent representation obtained from a demonstration of task $T_i$, 
and let $h^{(r)}(T_i)$ denote a reference embedding for the same task.
When learning a new task $T_k$, we penalize configurations where $g_t(T_k)$ is closer to old task references than to its own reference:

\begin{equation}
\label{eq:ara}
\begin{aligned}
\mathcal{L}_{\text{IFA}} 
= \frac{1}{|\mathcal{P}|}
\sum_{\substack{(j,k)\in\mathcal{P} \\ j<k}}
\max \Big( 0,\;
& d\big(g_t(T_k), h^{(r)}(T_k)\big) \\
& {} - d\big(g_t(T_k), h^{(r)}(T_j)\big) + \delta 
\Big) \: , 
\end{aligned}
\end{equation}

where $d(\cdot,\cdot)$ is a distance measure between two representations, $\mathcal{P}$ denotes the set of task pairs that are selected for adjustment, the loss is averaged over all pairs in $\mathcal{P}$, and $\delta$ controls the inter-task margin. 
The rationale of this adjustment is to encourage the representation of each new task to remain closer to its own reference embedding than to those of previously learned tasks. Accordingly, IFA introduces repulsive forces between the current task’s global latent representation and previous task references, while maintaining attractive forces toward its own reference. This mechanism promotes inter-task disentanglement and preserves within-task coherence, effectively mitigating representational interference.

If $\delta$ is fixed, it cannot adapt to each reference pair. The IFA loss may still take high values even when the current global latent representation is already sufficiently closer to the current task reference than to the previous task reference  (see Supp. for details).
To allow $\delta$ to adapt to the relative positions of references, we express the margin in terms of the distance between task references:

\begin{equation}
\label{eq:dist_delta}
\begin{aligned}
\delta &= \alpha\, d\!\left(h^{(r)}(T_k),\, h^{(r)}(T_j)\right)
\end{aligned} \: ,
\end{equation}

where $\alpha$ controls the margin. To magnify differences between close task references, instead of using the cosine distance, we propose to use the angular formulation of the distance:

\begin{equation}
d(a,b) = 
\arccos \!\left(
\frac{a^\top b}{\|a\|_2 \|b\|_2}
\right)
\end{equation}

In this way, our method can adapt to the distance between the task references.
Overall, during the lifelong stage, our training objective is:

\begin{equation}
\mathcal{L} =
\mathcal{L}_{\text{BC}}
+ \lambda_{\text{IFA}}\, \mathcal{L}_{\text{IFA}} \: .
\end{equation}

\subsection{Implementation Details} 
\label{sec:impDet}
\paragraph{Details for Base Policy.} 
Our policy (see Fig. \ref{fig:architecture}) consists of: vision, language, and state encoders; 
a FiLM layer that modulates visual and state features via language features; 
a temporal decoder; and a policy head. 
The inputs include \textit{agent-view} and \textit{eye-in-hand view} images (see definitions of such modalities in Sec. \ref{sec:experiments}) 
Specifically, we employ the image and text encoders from CLIP encoders~\cite{radford2021learning} 
as the vision and language encoders, respectively. 
A two-layer MLP serves as the state encoder, 
and a GPT-2 decoder~\cite{radford2019language}  is used as the temporal decoder. 
The vision and state embeddings are modulated by FiLM layers~\cite{perez2018film} 
to obtain task-conditioned representations.
{The rehearsal buffer $\mathcal{B}$ is populated by randomly sampling trajectories from the current task rollouts $\mathcal{T}_k$, while maintaining a balanced allocation across all previously encountered tasks.
}
 
\textbf{Task Pair Selection for IFA.}
Task pairs are drawn from the set $\mathcal{P}$ based on their similarity in both the language and agent-view modalities (see modalities definitions in Sec. \ref{sec:experiments}).
For each task, we extract the latent features $h^{(l)}$ (language) and $h^{(a)}$ (agent-view) from the current task data as well as from the previously stored samples in the buffer. 
For any two tasks $T_i$ and $T_j$, the modality-specific similarity is computed as the mean cosine similarity across all pairs of timesteps:
\vspace{-0.5em}
\begin{equation}
\text{Sim}_{e}(T_i, T_j) =
\frac{1}{N_i N_j} 
\sum_{t_1=1}^{N_i} \sum_{t_2=1}^{N_j}
\frac{ h^{(e)}_{t_1}(T_i)^{\top} h^{(e)}_{t_2}(T_j) }
     { \| h^{(e)}_{t_1}(T_i) \|_2 \, \| h^{(e)}_{t_2}(T_j) \|_2 }.
     \vspace{-0.2em}
\end{equation}

where $e \in \{\text{agent-view}, \text{language}\}$.
We compute $\text{Sim}_a$ and $\text{Sim}_l$ for all task pairs, rank them by similarity, and apply IFA only to pairs that (1) simultaneously fall within the top 50\% of most similar pairs in both modalities and (2) include one newly introduced and one previously learned task.

\textbf{Reference Choice.}
We take the language embedding $h^{(l)}$ of each task as its reference  $h^{(r)}(T_i)$  for IFA, being the task language description informative for the task at hand, and with a stable and fixed representation. We empirically justify our choice in the experimental section.  

%% file: sec/4_experiments.tex
\section{Experimental Analysis}
\label{sec:experiments}

\noindent \paragraph{Benchmark Suite and Datasets.}
We conduct all experiments using the LIBERO lifelong robotic manipulation benchmark~\cite{liu2023libero}.  
This benchmark provides a diverse set of manipulation tasks that closely resemble human daily activities, such as turning on a stove, moving books, or opening drawers.  
Each task is specified by a natural language instruction, for example: 
``\textit{Open the top drawer of the cabinet and put the bowl in it.}''
  
We follow the setup of prior SOTA, e.g.,~\cite{wan2024lotus,roy2025m2distill}, and adopt three task suites from the LIBERO benchmark: \textbf{LIBERO-OBJECT} (10 tasks), \textbf{LIBERO-GOAL} (10 tasks), and \textbf{LIBERO-50} (50 kitchen tasks selected from LIBERO-100). 
These suites respectively evaluate the robot’s ability to perform diverse actions, and handle complex long-horizon task compositions. All datasets incorporate visual, linguistic, and proprioceptive information, comprising \textit{agent-view} and \textit{eye-in-hand} cameras, textual task instructions, executed actions, and internal robot states, supporting multimodal policy learning. Specifically, \textit{agent-view} provides a visual perspective of the scene, offering spatial and contextual information about the environment, objects, and overall scene layout. \textit{Eye-in-hand} captures fine-grained visual information from a camera mounted on the robot’s end-effector. \textit{Language} encodes textual task instructions, guiding the policy by providing semantic goals and context.  \textit{Action} refers to the current control commands executed by the robot. Finally, \textit{state} encodes the robot’s proprioceptive information, including joint positions, velocities, and gripper status.

During the pretraining stage, six tasks are used for both LIBERO-OBJECT and LIBERO-GOAL ($|\mathcal{T}'| = 6$), and 25 tasks are used for LIBERO-50  ($|\mathcal{T}'| = 25$),  
with each pretraining task containing 50 demonstrations.  
In the lifelong learning stage, LIBERO-OBJECT and LIBERO-GOAL consist of four stages, with one new task introduced at each stage,  
while LIBERO-50 comprises five stages, with five new tasks added per stage to maintain computational feasibility. Each new task in the lifelong stage includes 10 demonstrations.

\begin{table*}[t!]
\centering
\caption{Comparison of results on the LIBERO benchmarks (mean $\pm$ standard deviation). ``NA'' indicates not available. Best results are shown in bold.
}
\label{tab:libero}
\resizebox{\textwidth}{!}{%
\begin{tabular}{lccc ccc ccc}
\toprule
\multirow{2}{*}{Method} &
\multicolumn{3}{c}{LIBERO-OBJECT} &
\multicolumn{3}{c}{LIBERO-GOAL} &
\multicolumn{3}{c}{LIBERO-50} \\
\cmidrule(lr){2-4}\cmidrule(lr){5-7}\cmidrule(lr){8-10}
 & FWT$\uparrow$ & NBT$\downarrow$ & AUC$\uparrow$
 & FWT$\uparrow$ & NBT$\downarrow$ & AUC$\uparrow$
 & FWT$\uparrow$ & NBT$\downarrow$ & AUC$\uparrow$ \\
\midrule
Sequential \cite{liu2023libero} & 62.0$\pm$1.0 & 63.0$\pm$2.0 & 30.0$\pm$1.0 & 55.0$\pm$1.0 & 70.0$\pm$1.0 & 23.0$\pm$1.0 & 32.0$\pm$1.0 & 90.0$\pm$2.0 & 14.0$\pm$2.0 \\
ER \cite{liu2023libero,chaudhry2019tiny} & 56.0$\pm$1.0 & 24.0$\pm$1.0 & 49.0$\pm$1.0 & 53.0$\pm$1.0 & 36.0$\pm$1.0 & 47.0$\pm$2.0 & 35.0$\pm$3.0 & 49.0$\pm$1.0 & 36.0$\pm$3.0 \\
BUDS \cite{zhu2022bottom,wan2024lotus} & 52.0$\pm$2.0 & 21.0$\pm$1.0 & 47.0$\pm$1.0 & 50.0$\pm$1.0 & 39.0$\pm$1.0 & 42.0$\pm$1.0 & 29$\pm$3.0 & 50.0$\pm$4.0 & 33.0$\pm$3.0 \\
LOTUS \cite{wan2024lotus} & 74.0$\pm$3.0 & \textbf{11.0$\pm$1.0} & 65.0$\pm$3.0 & 61.0$\pm$3.0 & 30.0$\pm$1.0 & 56.0$\pm$1.0 & 39$\pm$2.0 & 43.0$\pm$1.0 & 45.0$\pm$2.0 \\
ISCIL \cite{lee2024incremental} &  71.7$\pm$1.9 & 11.9$\pm$5.1 & 66.3$\pm$3.7 & 70.4$\pm$3.6 & 19.4$\pm$4.1 & 60.5$\pm$2.5 &  47.8$\pm$1.3 & 15.0$\pm$4.8 & 37.7$\pm$2.1  \\
M2Distill \cite{roy2025m2distill} &  75.0$\pm$3.0 & 8.0$\pm$5.0 & 69.0$\pm$4.0 & 71.0$\pm$1.0 & 20.0$\pm$3.0 & 57.0$\pm$2.0 &  NA & NA  & NA   \\
TAIL \cite{liu2023tail} & 41.5$\pm$2.9 & 22.5$\pm$2.1 & 38.2$\pm$2.5 &41.0$\pm$1.2 & 22.7$\pm$1.1 & 37.4$\pm$1.3 &  31.8$\pm$3.0  & 19.3$\pm$9.4  & 20.1$\pm$3.3  \\
\rowcolor{babyblue} MLR (Ours) & 83.3$\pm$2.6  &  12.3$\pm$2.4 & 77.6$\pm$3.0 & 78.8$\pm$2.2 & 10.0$\pm$7.5 & 74.6$\pm$2.7 &  58.0$\pm$4.0 & 20.6$\pm$9.0  & 54.7$\pm$2.4 \\
\rowcolor{babyblue} MLR + IFA (Ours) & \textbf{84.6$\pm$1.9} & 11.4$\pm$5.6 & \textbf{79.4$\pm$1.5} & \textbf{80.0$\pm$2.5} & \textbf{6.9$\pm$0.9} & \textbf{77.2$\pm$1.8} & \textbf{60.8$\pm$2.8} & \textbf{8.6$\pm$6.2} & \textbf{56.1$\pm$1.8} \\ 
\bottomrule
\end{tabular}
}
\end{table*}

\textbf{Evaluation Metrics.}
We evaluate policy performance using three standard metrics commonly employed in lifelong learning~\cite{wan2024lotus,rodriguez2018don,liu2023libero}:  
Forward Transfer (FWT), Negative Backward Transfer (NBT), and Area Under the Curve (AUC). All three metrics are computed in terms of success rates~\cite{liu2023libero}. While these metrics are widely used, their computation can slightly vary across studies; therefore, for consistency and fair comparison, we adopt the same definitions and calculation procedures as the studies we compare our method with, e.g., LOTUS~\cite{wan2024lotus}.

Let $M$ denote the total number of tasks in the lifelong learning sequence, and $r_{i,j}$ denote the agent’s success rate on task $j$ after it has learned the first $i$ tasks. FWT measures how well the agent adapts to new tasks and is defined as 
$\mathrm{FWT} = \tfrac{1}{M}\sum_{m=1}^{M} r_{m,m}$; higher FWT indicates faster adaptation.  
NBT measures forgetting on previous tasks, with lower values indicating less forgetting:  
$\mathrm{NBT}_m = \tfrac{1}{M-m}\sum_{q=m+1}^{M}(r_{m,m} - r_{q,m})$, and  
$\mathrm{NBT} = \tfrac{1}{M-1}\sum_{m=1}^{M-1} \mathrm{NBT}_m$.  
AUC reflects average performance across all tasks, computed as  
$\mathrm{AUC} = \tfrac{1}{M}\sum_{m=1}^{M} \mathrm{AUC}_m$, where  
$\mathrm{AUC}_m = \tfrac{1}{M-m+1}\bigl(r_{m,m} + \sum_{q=m+1}^{M} r_{q,m}\bigr)$; higher AUC indicates better overall success.

\textbf{Setup.}
Our experiments are repeated across three random seeds, with 20 trials per seed--each trial starting from a different configuration of the environment, i.e. different initial state of the gripper.
They are conducted on a single NVIDIA A100 GPU, with the pretraining stage and lifelong learning stage both using a batch size of 10 and trained for 100 epochs. We employed the AdamW optimizer~\cite{loshchilov2017decoupled} with a learning rate initialized to $10^{-4}$ and a linear scheduler.
The weighting coefficient $\lambda_{\text{IFA}}$ is fixed to 0.1. 
The best angular scaling factor $\alpha$ are 0.3, 0.7, and 0.1 for LIBERO-OBJECT, LIBERO-GOAL, and LIBERO-50, respectively (see Supp. Mat.).
We store the multimodal features in the buffer during MRL, with a total size equivalent to five demonstrations per task, following SOTA~\cite{wan2024lotus}.

\subsection{Results}
\label{sec:results} 

\subsubsection{Comparison with SOTA Methods} 
To assess the capability of our method to learn new tasks while mitigating forgetting of previous ones, we compare its performance against state-of-the-art methods (see Tab.~\ref{tab:libero}).
We include several different methods for comparison: Sequential, which fine-tunes each new task in order, using the ResNet–Transformer architecture proposed in~\cite{liu2023libero};
Experience Replay (ER)~\cite{liu2023libero,chaudhry2019tiny}, which stores raw trajectories in a buffer; 
BUDS, a hierarchical policy-based method~\cite{zhu2022bottom}, which was adapted in~\cite{wan2024lotus} to support LIL,
LOTUS~\cite{wan2024lotus}, ISCIL~\cite{lee2024incremental} and M2Distill~\cite{roy2025m2distill}; 
We also adapt TAIL~\cite{liu2023tail} to the LIL setting, following the protocol of LOTUS~\cite{wan2024lotus} (see Supp. Mat. for details).

The proposed MLR already yields substantial improvements over existing SOTA, such as LOTUS \cite{wan2024lotus}, ISCIL \cite{lee2024incremental}, and M2Distill \cite{roy2025m2distill}. In particular, MLR achieves higher FWT and AUC while maintaining competitive NBT values, confirming that our replay strategy alone effectively promotes knowledge retention and forward transfer.
When combined with IFA, the performance further improves on every metric. MLR + IFA achieves the highest FWT and AUC and among the lowest NBT values across all benchmarks. For example, on LIBERO-GOAL it improves AUC from 60.5 (ISCIL) to 77.2 while reducing NBT from 19.4 to 6.9, indicating a clear reduction in catastrophic forgetting. On the more challenging LIBERO-50 benchmark, MLR + IFA attains a large margin over all baselines, demonstrating superior scalability to longer and more diverse task sequences.
These results confirm that MLR provides an efficient and stable mechanism for continual imitation learning, while the IFA module further refines the replayed representations to enhance both stability and transfer. The combination delivers the best balance between knowledge reuse and plasticity, establishing a new SOTA on all LIBERO benchmarks.

\subsubsection{Ablation Study}
\label{sec:ablation}
We conducted a series of experiments to investigate several key aspects of the proposed method. In Tab.~\ref{tab:libero}, we compare our method using latent replay alone (MLR) and with the IFA component included (MLR + IFA). As seen, incorporating IFA improves performance across all datasets and metrics.
Below, we perform several ablations to examine the impact of different design factors, such as modality pairing, buffer configuration, and loss formulation, on the overall performance. Further results and analyses are included in the Supp. Mat. 

\begin{table}[t!]
\centering
\caption{
Influence of different modality similarity on the proposed method's performance. ``AV'', ``Lan'', ``EIH'', ``Act'', and ``Sta'' denote Agent-view, Language, Eye-in-hand, Action, and State modalities, respectively.
}
\label{tab:ablation_modality}
\resizebox{0.99\linewidth}{!}{
\begin{tabular}{lccccccccc}
\toprule
\textbf{Modality} & \multicolumn{3}{c}{\textbf{LIBERO-OBJECT}} & \multicolumn{3}{c}{\textbf{LIBERO-GOAL}} \\
\cmidrule(lr){2-4} \cmidrule(lr){5-7}
 & FWT$\uparrow$ & NBT$\downarrow$ & AUC$\uparrow$ & FWT$\uparrow$ & NBT$\downarrow$ & AUC$\uparrow$ \\
\midrule
AV & 83.8$\pm$2.2 & \textbf{11.1$\pm$10.4} & 78.9$\pm$2.6 & 77.7$\pm$2.6  & 7.8$\pm$2.7 & 73.9$\pm$2.6 \\
Lan & 83.3$\pm$0.7  &  13.5$\pm$3.1 & 77.3$\pm$1.9 & 77.1$\pm$1.9 & 12.3$\pm$0.8 & 71.8$\pm$2.0  \\
EIH & 84.4$\pm$0.9  &  16.1$\pm$3.1 & 77.1$\pm$1.1 & 75.0$\pm$3.3  &  5.6$\pm$3.9 & 72.7$\pm$4.2  \\
Act & \textbf{84.6$\pm$2.6}  &  20.7$\pm$3.6 & 75.1$\pm$1.4 & 73.8$\pm$2.5 & \textbf{2.5$\pm$4.2} & 73.2$\pm$1.1  \\
Sta & 82.5$\pm$3.3  &  11.2$\pm$10.6 & 77.5$\pm$2.6 & 78.3$\pm$5.1 & 8.2$\pm$2.4 & 74.9$\pm$5.9  \\
Lan + Act & 82.5$\pm$3.3  &  11.2$\pm$10.6 & 77.5$\pm$2.6 & 75.0$\pm$3.3  &  5.6$\pm$3.9 & 72.7$\pm$4.2 \\ 
Lan + EIH & 83.3$\pm$1.4  &  16.5$\pm$9.0 & 75.6$\pm$5.3 & 75.0$\pm$3.3  &  5.6$\pm$3.9 & 72.7$\pm$4.2 \\      
Lan + Sta & \textbf{84.6$\pm$2.6}  &  17.4$\pm$6.4 & 76.7$\pm$0.6 & 75.0$\pm$3.3  &  5.6$\pm$3.9 & 72.7$\pm$4.2 \\
 Lan + AV & \textbf{84.6$\pm$1.9} & 11.4$\pm$5.6 & \textbf{79.4$\pm$1.5} & \textbf{80.0$\pm$2.5} & 6.9$\pm$0.9 & \textbf{77.2$\pm$1.8} \\
\bottomrule
\end{tabular}}
\end{table}

\begin{table}[t!]
\centering
\caption{Ablation on task pair selection proportion used when calculating IFA.}
\label{tab:ablation_prop}
\resizebox{\linewidth}{!}{
\begin{tabular}{llccc}
\toprule
\textbf{} & \textbf{Buffer} & \textbf{FWT}$\uparrow$ & \textbf{NBT}$\downarrow$ & \textbf{AUC}$\uparrow$ \\
\midrule
\multirow{3}{*}{LIBERO-OBJECT}
  & 33.3\% & 83.3$\pm$2.6  &  12.3$\pm$2.4 & 77.6$\pm$3.0 \\
  &  50\%   & \textbf{84.6$\pm$1.9} & \textbf{11.4$\pm$5.6} & \textbf{79.4$\pm$1.5} \\
  & 66.6\% & 83.3$\pm$1.4 & 16.4$\pm$9.0 & 75.8$\pm$5.3 \\
\midrule
\multirow{3}{*}{LIBERO-GOAL}
  & 33.3\% & 80.0$\pm$2.5 & \textbf{6.9$\pm$0.9} & \textbf{77.2$\pm$1.8} \\
  &  50\%   & 80.0$\pm$2.5 & \textbf{6.9$\pm$0.9} & \textbf{77.2$\pm$1.8} \\
  & 66.6\% & \textbf{82.0$\pm$0.9} & 15.4$\pm$16.7 & 75.6$\pm$8.7 \\
\bottomrule
\end{tabular}
}
\end{table}

\textbf{Influence of Different Modality Similarity on IFA.}
Based on the task pair selection for IFA described in Sec.~\ref{sec:impDet}, we analyze how using an alternative modality $e$ to compute 
$\text{Sim}_{e}(T_i, T_j)$ affects the effectiveness of IFA and, consequently, the overall performance of the proposed method. 
In addition to evaluating each modality individually, we also consider combinations with language, since language serves as the reference in our design.
As shown in Tab.~\ref{tab:ablation_modality}, the language and agent-view pairing (that is the one our approach leverages) achieves the best overall performance, improving both AUC and FWT while maintaining low forgetting (NBT). \\


\begin{table}[t!]
\centering
\caption{
Effectiveness of choosing different references. ``Lan'' stands for language. See the text for a description of the ``mean global'' reference.
}
\label{tab:ablation_ reference}
\setlength{\tabcolsep}{8pt}
\renewcommand{\arraystretch}{1.15}
\resizebox{\linewidth}{!}{
\begin{tabular}{llccc}
\toprule
\textbf{} & \textbf{Reference} & \textbf{FWT}$\uparrow$ & \textbf{NBT}$\downarrow$ & \textbf{AUC}$\uparrow$ \\
\midrule
\multirow{2}{*}{LIBERO-OBJECT}
  & Mean global & 79.6$\pm$7.21 & \textbf{8.3$\pm$12.2} & 75.7$\pm$1.1 \\
    &   Lan+agent view & \textbf{84.6$\pm$1.9} & 11.4$\pm$5.6 & \textbf{79.4$\pm$1.5} \\
  \midrule
\multirow{2}{*}{LIBERO-GOAL}
  & Mean global & 77.1$\pm$3.1  &  15.7$\pm$6.1 & 70.5$\pm$6.0 \\
   &  Lan+agent view & \textbf{80.0$\pm$2.5} & \textbf{6.9$\pm$0.9} & \textbf{77.2$\pm$1.8}  \\
\bottomrule
\end{tabular}}
\end{table}

\begin{table}[t!]
\centering
\caption{Impact of varying the probability of storing features in the buffer, affecting buffer size.}
\label{tab:ablation_buffer}
\resizebox{\linewidth}{!}{
\begin{tabular}{lccccc}
\toprule
\textbf{} & \textbf{Probability} & \textbf{Storage} & \textbf{FWT}$\uparrow$ & \textbf{NBT}$\downarrow$ & \textbf{AUC}$\uparrow$ \\
\midrule
\multirow{3}{*}{LIBERO-OBJECT}
  & 0.50 & 188MB & 84.6$\pm$1.9 & 11.4$\pm$5.6 & 79.4$\pm$1.5 \\
  & 0.20 & 75.3MB & 80.5$\pm$7.5 & 12.5$\pm$7.1 & 77.7$\pm$10.0 \\
  & 0.10 & 37.6MB & 79.9$\pm$12.6 & 13.4$\pm$6.1 & 76.6$\pm$14.1 \\

\midrule
\multirow{3}{*}{LIBERO-GOAL}
  & 0.50 & 121.2MB & 80.0$\pm$2.5 & 6.9$\pm$0.9 & 77.2$\pm$1.8 \\
  & 0.20 & 60.6MB & 78.3$\pm$1.9 & 10.6$\pm$8.0 & 77.4$\pm$4.6 \\
  & 0.10 & 30.3MB & 78.7$\pm$1.3 & 25.2$\pm$10.4 & 66.2$\pm$6.8 \\
\bottomrule
\end{tabular}}
\end{table}

\textbf{Effect of Task Pair Selection Proportion.}
After identifying the language and agent-view modalities as the criteria for IFA as shown in Tab.~\ref{tab:ablation_modality}, we further investigate how varying the proportion of selected task pairs influences the performance.  
The experimental setup is consistent with the previous section,  
with the only difference being that the independent variable is the selection proportion of task pairs.
As shown in Tab.~\ref{tab:ablation_prop}, selecting the top 50\% of task pairs consistently achieves the highest FWT and AUC while maintaining low NBT, indicating the most stable and effective overall performance. \\


\textbf{Effectiveness of Choosing Different References.}
Different from using the latent feature of language as the reference, i.e., $h^{(r)}(T_k) = h^{(l)}(T_k)$, we also consider the centroid of each task’s global latent representations. For previous tasks, we computed the mean of all the global latent representations for each task.
During the lifelong learning of the next stage, this means representation serves as the reference for the corresponding previous task. For the new task, we use the mean of its global latent representations accumulated from the start of the current epoch as its reference. All other settings remain unchanged.
As shown in Tab.~\ref{tab:ablation_ reference}, using the latent feature of language as the reference consistently achieves better performance than the averaged global representation. \\


\textbf{Impact of Varying Data Storage Ratios.}
We investigate the impact of varying the buffer size on lifelong learning performance, by setting the probability of storing features in the buffer. In particular, we use either 0.10, 0.20, or 0.50 probability of storing a feature for each new task demonstration. As shown in Tab.~\ref{tab:ablation_buffer}, retaining fewer representations generally leads to performance degradation, while storing a larger fraction consistently yields more stable results across tasks. \\


\begin{figure}
    \centering
    \includegraphics[width=1\linewidth]{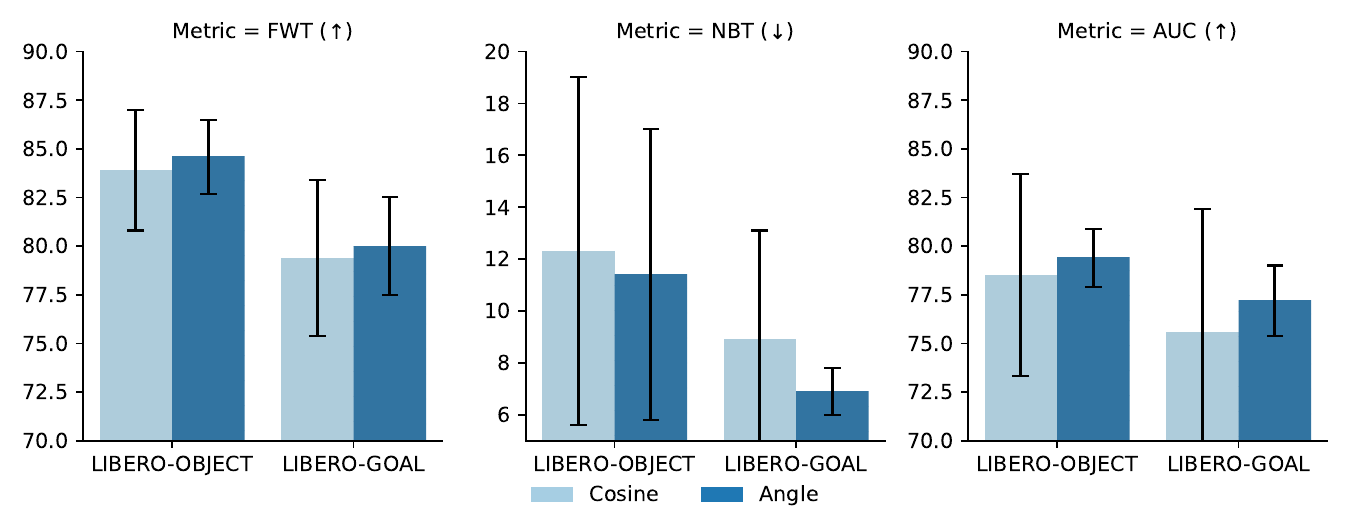}
    \caption{Comparison between cosine distance and angle-based IFA loss calculation.}
    \label{tab:ara_cos_vs_angle}
\end{figure}

\textbf{Angle-Based versus Cosine Distance-Based Loss.}
We further compare the angle-based loss described in Sec. \ref{sec:ourapproach} with the cosine distance loss.  
As shown in Fig.~\ref{tab:ara_cos_vs_angle}, the angle-based loss calculation consistently outperforms the cosine distance-based calculation across all metrics. Notably, even the cosine distance–based calculation surpasses all SOTA methods and improves upon the MLR-only setting (see Tab.~\ref{tab:libero}). However, its standard deviation is notably higher. \\

\begin{figure}[t!]
    \centering
    \begin{subfigure}[b]{0.495\linewidth}
        \centering
        \includegraphics[width=\textwidth]{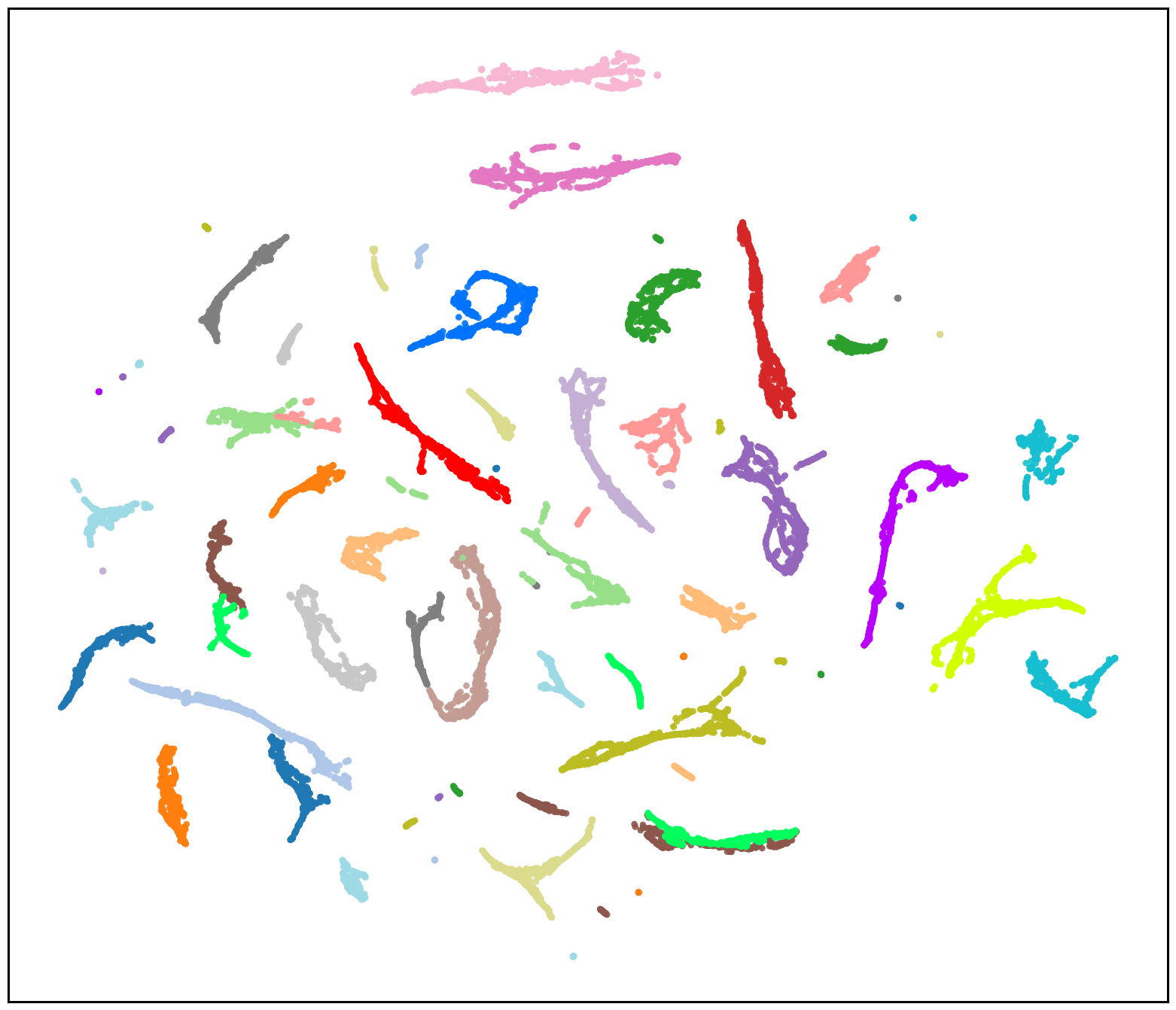}
        \caption{w/o IFA.}
        \label{fig:umap_before}
    \end{subfigure}
    \hfill
    \begin{subfigure}[b]{0.495\linewidth}
        \centering
        \includegraphics[width=\textwidth]{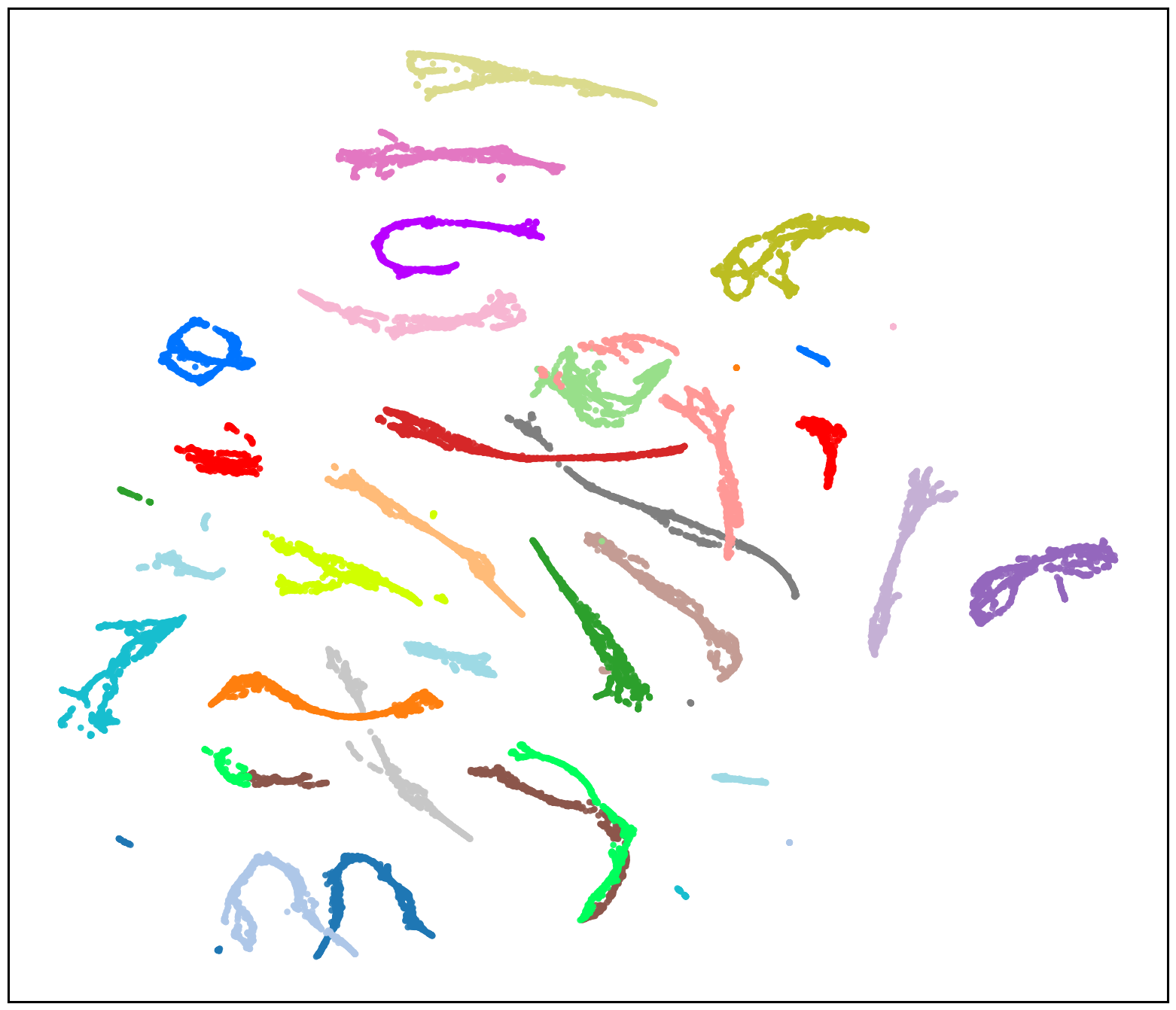}
        \caption{w/ IFA.}
        \label{fig:umap_after}
    \end{subfigure}
    \caption{UMAP visualization of the global latent representation $g(T_k)$ obtained when not enforcing the IFA loss (left) or when enforcing it (right). Each color represents the global latent representations of one task.}
    \label{fig:ifa_umap}
\end{figure}

\begin{table}[t!]
\centering
\caption{Average cosine similarity between task~8 embeddings and references from other tasks after lifelong training. ``w/o'' and ``w'' denote training without and with IFA, respectively (see main text for details).}
\label{tab:dis}
\small
\resizebox{\linewidth}{!}{
\renewcommand{\arraystretch}{1.2} 
\begin{tabular}{l ccc ccc}
\toprule
\textbf{} &  \multicolumn{3}{c}{\textbf{LIBERO-OBJECT}} &  \multicolumn{3}{c}{\textbf{LIBERO-GOAL}} \\
\cmidrule(lr){2-4} \cmidrule(lr){5-7}
& T6 & T7 & T8 & T6 & T7 & T8 \\ 
\midrule
w/o & 0.1265 & 0.1374 & 0.3590 & 0.2191 & 0.2319 & 0.3652 \\
w   & 0.1197 & 0.1218 & 0.3482 & 0.1977 & 0.2345 & 0.3877 \\
\bottomrule
\end{tabular}}
\end{table}

\textbf{IFA effects in the latent space.}
To better highlight the advantages brought by the proposed angle-based IFA in the learning dynamics, we compare the global latent representation distributions obtained both training without applying IFA (denoted with ``w/o'') or applying it (denoted with ``w''), with all other settings identical, using UMAP~\cite{mcinnes2018umap} visualization, shown in Fig.~\ref{fig:ifa_umap}.
Notice that all experimental settings are identical except for the presence of IFA. Without the contribution coming from the IFA loss (Fig.~\ref{fig:ifa_umap}-left), embeddings of different tasks are intertwined, showing low inter-task separation and moderate intra-task cohesion, which can lead to catastrophic forgetting as new task features may overwrite old ones. 
When the IFA component is enforced (Fig.~\ref{fig:ifa_umap}-right), embeddings form compact, distinct clusters around their task-specific references, reflecting higher inter-task separation and improved intra-task cohesion.
This aligns with the IFA loss, which penalizes embeddings closer to other tasks’ references than their own, enforcing repulsion between tasks and attraction to their own reference. The bottom plot, together with the quantitative results presented in this paper, confirms that IFA successfully produces task-aware embeddings, reducing interference and stabilizing lifelong learning by organizing the feature space into distinct, task-specific clusters. Notably, Fig.~\ref{fig:ifa_umap} also shows some task embeddings are already
separated without IFA. In this situation, IFA targets such high-similarity 
cases using an adaptive, angle-based margin, which preserves small distinctions that cosine similarity compresses. Therefore, very similar tasks might still remain highly similar or overlapping. 

To further examine IFA's effect on task-level feature separation, we report in Tab.~\ref{tab:dis} the average cosine similarity between task~8 and references from other tasks after lifelong training. Higher cosine similarity indicates closer alignment in the reference space, which can lead to interference between tasks. On LIBERO-OBJECT, where IFA is only applied between tasks~8 and~7 (i.e., no IFA used for any other task pairs), the similarity between task~8 and task~7 decreases from 0.1374 to 0.1218. Similarly, on LIBERO-GOAL, where IFA is only applied between tasks~8 and~6, the similarity between task~8 and task~6 decreases from 0.2191 to 0.1977. Such consistent reductions indicate that IFA successfully increases inter-task separability as intended. 

%% file: sec/5_conclusion.tex
\section{Conclusion}
We presented a lifelong imitation learning framework that enables agents to continuously acquire new skills while mitigating catastrophic forgetting. By having our Multimodal Latent Replay and Incremental Feature Adjustment, our method efficiently stores and reuses compact multimodal representations, leveraging frozen pretrained encoders to achieve robust performance even when tasks share similar latent features. Our approach sets a new SOTA on LIBERO benchmarks, improving task transfer and reducing forgetting across diverse multimodal tasks. Looking forward, exploring more complex or longer task sequences, cross-domain and real-world robotic scenarios, and integration with reinforcement learning offers exciting opportunities to further extend the adaptability and generality of our framework, addressing challenges that are broadly difficult in lifelong learning research.

%% file: sec/X_suppl.tex











\appendix

\clearpage
\setcounter{page}{1}
\maketitlesupplementary

This supplementary material includes extended implementation details of our method, as well as descriptions of the state-of-the-art methods used for comparison in Sec.~\ref{SUPMATsec:addition_para}. Next, we present additional ablation studies that reflect and justify the design choices of our components and their parameter configurations in Sec.~\ref{SUPPMATsec:ab}.  Sec.~\ref{SUPPsec:extended} offers further analyses on IFA and the choice of the task reference embedding, including an explanation of why angle-based computation can be more effective and how the adaptive parameter is designed. Finally, we report the computational efficiency of our method in Sec.~\ref{SUPPMATsec:comp}.

\section{Extended Implementation Details}
\label{SUPMATsec:addition_para}
\subsection{Our Framework}
During multi-task pre-training, each task $\mathcal{T}'$ includes $N_j = 50$ expert demonstrations. 
In the lifelong learning stage, each new task $\mathcal{T}_k$ provides $N_k = 10$ demonstrations, following the setup used in prior state-of-the-art (SOTA) methods \cite{wan2024lotus, roy2025m2distill}. 
Each rollout in the LIBERO suites consists of 70--300 time steps.

Our architecture follows the general design introduced in prior work~\cite{liu2023tail}, where each action is predicted from the previous $L=8$ observations. Specifically, we employ CLIP-base~\cite{radford2021learning} as both the vision encoder and the language encoder, each implemented as a 12-layer Transformer. A 6-layer GPT-2 decoder~\cite{radford2019language} serves as the temporal decoder, processing the visual and state tokens from the last eight steps and outputting the parameters of a 5-component Gaussian Mixture Model (GMM) over continuous actions. The use of a GMM policy head is a common practice in the LIBERO benchmarks~\cite{liu2023libero}. Task instructions are injected via FiLM-based~\cite{perez2018film} conditioning of the visual and state representations within the modulation network. During pre-training, we train only LoRA adapters (with rank-8) in the CLIP backbones, together with the temporal decoder and the GMM policy head. In the lifelong learning stage, only the temporal decoder and policy head are updated for each new task, as depicted in Figure \ref{fig:architecture} in the main paper.


\subsection{State of the Art}

For the methods reported in Tab.~1 of the main paper, Sequential~\cite{liu2023libero}, Experience Replay (ER)~\cite{liu2023libero,chaudhry2019tiny}, BUDS~\cite{zhu2022bottom,wan2024lotus}, and LOTUS~\cite{wan2024lotus} all use the same architectural setup and training configuration as LOTUS. Specifically, LOTUS employs a skill-discovery module to predict subgoal actions, which are then composed into a full action sequence to accomplish the task, and the numerical results for these methods are taken directly from the original LOTUS paper~\cite{wan2024lotus}.
As for ISCIL~\cite{lee2024incremental}, each skill is associated with its own adapter, and during evaluation the method retrieves the adapter corresponding to the skill most similar to the current input. 
For ISCIL, we use the implementation from~\cite{lei2025dynamic}, which has been adapted for the LIBERO setting. In this way, the same training setup as other methods is followed, ensuring fair comparisons.

We exploited an in-house implementation of TAIL~\cite{liu2023tail} (note that the official code for TAIL~\cite{liu2023tail} is not available). We verified that our implementation has the same number of parameters as reported in the TAIL paper, and therefore use it for all comparisons with confidence. As the official implementation is unavailable, we ensured architectural and training consistency with the original paper to maintain fair comparisons. We will make the corresponding code publicly available upon acceptance of our paper to foster future research in this direction.

TAIL~\cite{liu2023tail} is a task-ID–based method that trains a separate adapter for each task to adapt the shared backbone to the corresponding data distribution. At test time, TAIL requires the task ID to select the correct adapter. To make TAIL compatible with the task-ID–agnostic LIL setting considered in the main paper, we first train it following the original procedure, without any modification.
During evaluation, since multiple task-specific adapters are available at a given stage of the lifelong phase, we compute the success rate of each task as the average performance across all adapters available up to that point. This averaging simulates task-ID agnostic inference, where the correct adapter cannot be selected, and follows the same evaluation strategy adopted in LOTUS~\cite{wan2024lotus} when adapting task-incremental methods to LIL. All reported metrics use this averaged success rate, ensuring a fair comparison with LIL approaches that employ a single shared policy.

\section{Additional Ablation Studies}
\label{SUPPMATsec:ab}

\subsection{Influence of the Angular Scaling Parameter 
\texorpdfstring{$\alpha$}{alpha} in IFA}
  
We conducted an ablation study to evaluate the impact of different values of the angular scaling factor $\alpha$ in the adaptive IFA formulation (Tab.~\ref{tab:ablation_m}). The best-performing values of $\alpha$ are 0.3, 0.7, and 0.1 for LIBERO-OBJECT, LIBERO-GOAL, and LIBERO-50, respectively.

The adaptive margin $\delta$ is defined as proportional to the angular distance between the reference embeddings of the current task and the most similar previous task: $\delta = \alpha  \arccos\big( \frac{a^\top b}{|a|_2 |b|_2} \big)$. Intuitively, this ensures that tasks with larger semantic differences are more strongly separated in the latent space, while similar tasks remain closer, preserving within-task coherence.

For LIBERO-50, the largest and most diverse benchmark, smaller $\alpha$ (0.1) gives the best overall performance, likely reflecting the denser arrangement of task reference embeddings—larger margins could over-penalize nearby tasks. However, performance at $\alpha$ = 0.7 remains comparable, indicating that the adaptive IFA is relatively robust to the choice of $\alpha$ even in large, densely packed task suites. This suggests that $\alpha$ can be tuned to balance inter-task separation and latent stability without critically affecting performance.

\begin{table}[t!]
\centering
\caption{Ablation on  different angular scaling factor $\alpha$ in the adaptive IFA.}
\label{tab:ablation_m}
\setlength{\tabcolsep}{8pt}
\renewcommand{\arraystretch}{1.15}
\resizebox{0.95\linewidth}{!}{
\begin{tabular}{llccc}
\toprule
\textbf{Dataset} & \textbf{ $\alpha$} & \textbf{FWT}$\uparrow$ & \textbf{NBT}$\downarrow$ & \textbf{AUC}$\uparrow$ \\
\midrule
\multirow{4}{*}{LIBERO-OBJECT}
  & 0.10 & 81.8$\pm$3.2 & 12.4$\pm$3.6 & 77.3$\pm$1.4 \\
  & 0.30 & \textbf{84.6$\pm$1.9} & 11.4$\pm$5.6 & \textbf{79.4$\pm$1.5} \\
  & 0.50 & 82.1$\pm$0.7 & 18.3$\pm$10.3 & 73.2$\pm$4.1 \\
  & 0.70 & 81.7$\pm$0.7 & \textbf{8.4$\pm$3.3} & 77.3$\pm$2.4 \\

\midrule
\multirow{4}{*}{LIBERO-GOAL}
  & 0.10 & \textbf{80.8$\pm$4.0} & 12.0$\pm$11.3 & 75.4$\pm$8.8 \\
  & 0.30 & 81.3$\pm$3.3 & 15.2$\pm$7.3 & 74.7$\pm$6.7 \\
  & 0.50 & 79.2$\pm$1.9 & 12.8$\pm$2.8 & 73.1$\pm$2.3 \\
  & 0.70 & 80.0$\pm$2.5 & \textbf{6.9$\pm$0.9} & \textbf{77.2$\pm$1.8} \\

\midrule
\multirow{4}{*}{LIBERO-50}
  & 0.10 & \textbf{60.8$\pm$2.8}& 8.6$\pm$6.2 & \textbf{56.1$\pm$1.8}\\
  & 0.30 & 53.6$\pm$4.0 & \textbf{7.6$\pm$10.0} & 52.0$\pm$1.4  \\
  & 0.50 & 58.5$\pm$3.6 & 8.9$\pm$4.9 & 53.2$\pm$3.4  \\
  & 0.70 & 60.5$\pm$2.2 & 10.6$\pm$1.1 & 54.2$\pm$2.3 \\

\bottomrule
\end{tabular}}
\end{table}

\begin{table}[t!]
\centering
\caption{Comparison between LoRA adapters with different ranks (R) and full fine-tuning (FFT). 
}
\label{tab:lora}
\small
\resizebox{\linewidth}{!}{
\renewcommand{\arraystretch}{1.2} 
\begin{tabular}{l ccc ccc}
\toprule
\textbf{} &  \multicolumn{3}{c}{\textbf{LIBERO-OBJECT}} &  \multicolumn{3}{c}{\textbf{LIBERO-GOAL}} \\
\cmidrule(lr){2-4} \cmidrule(lr){5-7}
& \textbf{FWT}$\uparrow$ & \textbf{NBT}$\downarrow$ & \textbf{AUC}$\uparrow$ & \textbf{FWT}$\uparrow$ & \textbf{NBT}$\downarrow$ & \textbf{AUC}$\uparrow$ \\ 
\midrule
R=8 & 57.5$\pm$0.2 & 7.9$\pm$1.0 & 52.8$\pm$5.9 & 59.3$\pm$0.9 & 11.5$\pm$3.3 & 54.1$\pm$0.7  \\
R=16 & 48.9$\pm$7.1 & \textbf{-5.6$\pm$0.4} & 50.4$\pm$6.9 & 56.3$\pm$5.3 & \textbf{4.4$\pm$18.9} & 53.6$\pm$2.4 \\
R=32 &  53.8$\pm$0.2 & -0.2$\pm$6.7 & 54.2$\pm$2.3 & 63.8$\pm$7.1 & 7.5$\pm$13.0 & 60.0$\pm$1.5   \\
FFT & \textbf{84.6$\pm$1.9} & 11.4$\pm$5.6 & \textbf{79.4$\pm$1.5} & \textbf{80.0$\pm$2.5} & 6.9$\pm$0.9 & \textbf{77.2$\pm$1.8} \\

\bottomrule
\end{tabular}}
\end{table}

\subsection{Adapters versus Full Fine-tuning.}

During the lifelong learning stage, our default configuration fine-tunes all parameters of the temporal decoder. For comparison, we also evaluate a parameter-efficient variant based on Low-Rank Adaptation (LoRA) \cite{hu2022lora}, where the temporal decoder is frozen and LoRA modules are inserted into the attention projection layers (query/key/value and output projections) and the MLP feed-forward layers. We experiment with different LoRA ranks, and the results are summarized in Tab.~\ref{tab:lora}.

These results reveal a substantial performance gap: full fine-tuning significantly outperforms the LoRA-based adaptation across all datasets and metrics. This indicates that, in our setting, the temporal decoder requires sufficient capacity to capture complex temporal dynamics and integrate information coming from the MLR and IFA losses, which parameter-efficient approaches alone cannot provide. Furthermore, LoRA introduces additional hyperparameters that require grid search, making it less practical for large-scale lifelong learning scenarios.

\begin{table}[t!]
\centering
\caption{Effect of using FiLM layers during lifelong learning. 
}
\label{tab:film}
\small
\resizebox{\linewidth}{!}{
\renewcommand{\arraystretch}{1.2} 
\begin{tabular}{l ccc ccc}
\toprule
\textbf{} &  \multicolumn{3}{c}{\textbf{LIBERO-OBJECT}} &  \multicolumn{3}{c}{\textbf{LIBERO-GOAL}} \\
\cmidrule(lr){2-4} \cmidrule(lr){5-7}
& \textbf{FWT}$\uparrow$ & \textbf{NBT}$\downarrow$ & \textbf{AUC}$\uparrow$ & \textbf{FWT}$\uparrow$ & \textbf{NBT}$\downarrow$ & \textbf{AUC}$\uparrow$ \\ 
\midrule
w/o FiLM  & 43.8$\pm$3.5 & \textbf{8.8$\pm$7.7} & 41.6$\pm$0.2 & 57.6$\pm$17.7 & \textbf{0.83$\pm$7.5} & 56.3$\pm$2.1 \\
w FiLM  & \textbf{84.6$\pm$1.9} & 11.4$\pm$5.6 & \textbf{79.4$\pm$1.5} & \textbf{80.0$\pm$2.5} & 6.9$\pm$0.9 & \textbf{77.2$\pm$1.8} \\
\bottomrule
\end{tabular}}
\end{table}

\subsection{Influence of Using FiLM Layers.}

In our model, the outputs of the vision and state encoders are modulated by FiLM layers~\cite{perez2018film} to obtain task-conditioned representations. To empirically justify their contribution, we ablate the model by removing FiLM layers during the lifelong learning stage (see Tab.~\ref{tab:film}).

As seen, incorporating FiLM layers significantly improves FWT and AUC compared to the variant without FiLM. Although the no-FiLM variant shows slightly lower NBT, this is due to its overall weaker learning performance rather than genuine resistance to forgetting. These results indicate that explicit feature-wise modulation is crucial for effective adaptation: by dynamically scaling and shifting intermediate activations, FiLM allows the network to adjust its internal representations efficiently to the conditioning signal, improving forward transfer and overall learning effectiveness.


\begin{table*}[t]
\centering
\caption{
Task-reference comparison on \textbf{LIBERO-OBJECT} and \textbf{LIBERO-GOAL}.
See text for the explanation of ``Mean Global'' and 
``Global''.
``Language'', ``AgentView'', ``Eye-in-hand'', and ``State'' use the corresponding modality-specific latent feature as the task reference. The smallest cosine distance is highlighted in \textbf{bold}, and the second smallest is \underline{underlined}.
}
\label{tab:modality_similarity}
\small
\renewcommand{\arraystretch}{1.1}
\setlength{\tabcolsep}{4pt}
\begin{tabular*}{\linewidth}{@{\extracolsep{\fill}}lcccc|cccc}
\toprule
 & \multicolumn{4}{c|}{\textbf{LIBERO-OBJECT}} 
 & \multicolumn{4}{c}{\textbf{LIBERO-GOAL}} \\
\cmidrule(lr){2-5}\cmidrule(lr){6-9}
\textbf{Task Reference} 
& \textbf{Task 6} & \textbf{Task 7} & \textbf{Task 8} & \textbf{Task 9}
& \textbf{Task 6} & \textbf{Task 7} & \textbf{Task 8} & \textbf{Task 9} \\
\midrule
Mean global    
& \textbf{0.565} & \textbf{0.555} & \textbf{0.549} & \textbf{0.454}
& \textbf{0.485} & \textbf{0.354} & \textbf{0.501} & \textbf{0.437} \\
Global       
& 0.811  & 0.802  & 0.797  & 0.703 
& 0.735  & \underline{0.583} & 0.752  & 0.684 \\
Language  
& \underline{0.668} & \underline{0.701} & \underline{0.656} & \underline{0.677}
& \underline{0.591} & 0.610 & \underline{0.629} & \underline{0.680} \\
AgentView 
& 0.947 & 0.951 & 0.973 & 0.994
& 0.976 & 0.957 & 0.980 & 1.051 \\
Eye-in-hand 
& 1.008 & 0.974 & 1.005 & 1.005
& 1.018 & 1.037 & 0.987 & 1.042 \\
State     
& 1.068 & 1.012 & 1.024 & 1.006
& 1.033 & 1.072 & 1.005 & 1.036 \\
\bottomrule
\end{tabular*}
\end{table*}

\begin{figure}[t!]
    \centering
   \includegraphics[width=0.395\textwidth]{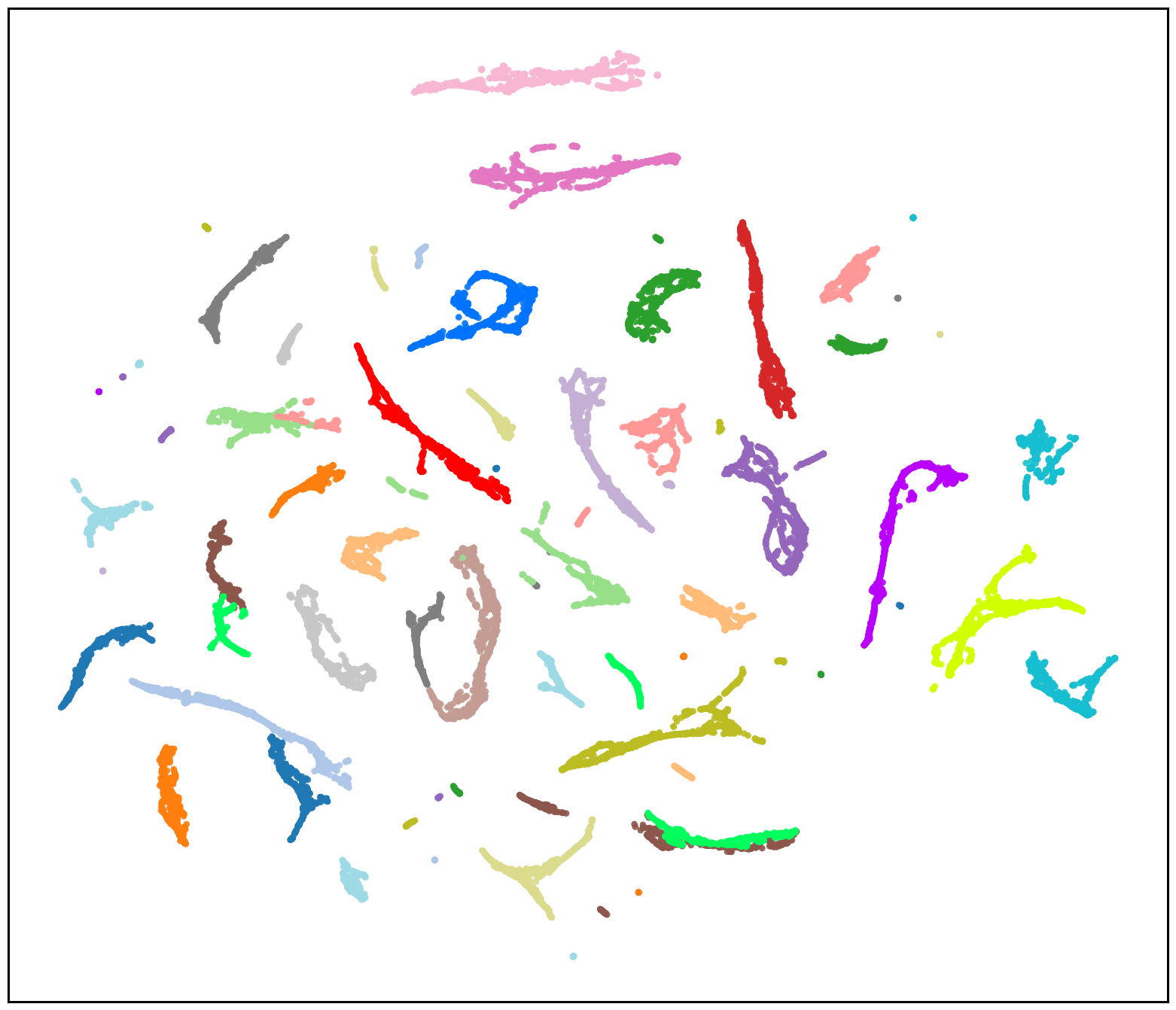}
        
    \caption{
    UMAP visualization of the global latent representations $g(T_k)$ obtained when not enforcing the IFA loss, each color represents the global latent representations of one task.}
    \label{fig:ifa_umap1}
\end{figure}

\section{Extended Analysis}
\label{SUPPsec:extended}
\subsection{Selection of a stable task reference}

As described in the main paper, our proposed Incremental Feature Adjustment (IFA) is a regularization technique designed to counter representation drift across tasks during Lifelong Learning (LIL).
IFA achieves this by ensuring that the global latent representation ($g_t(T_k)$), which is the output of the temporal decoder for the current task $T_k$, remains closer to its own task reference ($h^{(r)}(T_k)$) than to the references of previously learned tasks ($h^{(r)}(T_j)$). This introduces repulsive forces between the current task’s global latent representation and previous task references, promoting inter-task disentanglement and preserving within-task coherence.

The efficacy of IFA relies critically on selecting a stable and representative task reference ($h^{(r)}$) for each task. The motivation for this selection can be seen from the visualization in $\text{Fig.}~\ref{fig:ifa_umap1}$: we observe that the global latent representations of a given task naturally cluster together in the embedding space. This clustering indicates that each task can be summarized by a single, representative point that reflects the overall location of its features.

To justify our choice of the language latent feature as the task reference ($h^{(r)}$), we compared it against other viable candidates. We computed the average cosine distance between these candidates and the global latent representations of their corresponding tasks (on LIBERO-OBJECT and LIBERO-GOAL, Tasks 6–9), evaluating both their representativeness and their stability during incremental learning.
The task reference candidates evaluated were the mean latent feature derived from different modalities (first column of Tab.~\ref{tab:modality_similarity}): Mean global, the mean of all global latent representations for a task; Global, a measure derived by averaging the distances obtained when treating each individual global latent representation as a separate, unstable candidate reference; 
Modality-Specific Latent Features, the mean latent feature of a specific modality, such as language, agent view, eye-in-hand, or state (see Section \ref{sec:experiments}). For example, when using the agent-view modality, we take the mean agent-view latent features of the task as the task reference and compute its average cosine distance to all global latent representations of that task (see Tab.~\ref{tab:modality_similarity}). This data confirmed that using the mean of all global latent representations (the Global Latent Feature candidate) yields the smallest cosine distance, proving it is the most statistically representative point.
Despite being the most representative, the mean global reference is fundamentally unstable. During the ongoing process of Lifelong Incremental Learning (LIL), the dynamic changes in model parameters cause the latent representations, and consequently their mean, to change over time. This drifting reference hinders the ability of IFA to consistently regulate representation drift.
Crucially, the language latent feature produced the second-smallest distances. Furthermore, the language-based task reference (which is identical to its mean since the language modality provides a single, fixed embedding per task description) remains stable throughout training. It even outperformed the unstable "Global" baseline. Given that a stable anchor is essential for IFA to effectively prevent representation drift across tasks, we adopt the language latent feature as our definitive task reference ($h^{(r)}$), prioritizing its stability for reliable regularization over the slightly higher representativeness of the unstable mean global feature.

\begin{figure}[t!]
    \centering
    \begin{subfigure}[b]{0.495\linewidth}
        \centering
        \includegraphics[width=\textwidth]{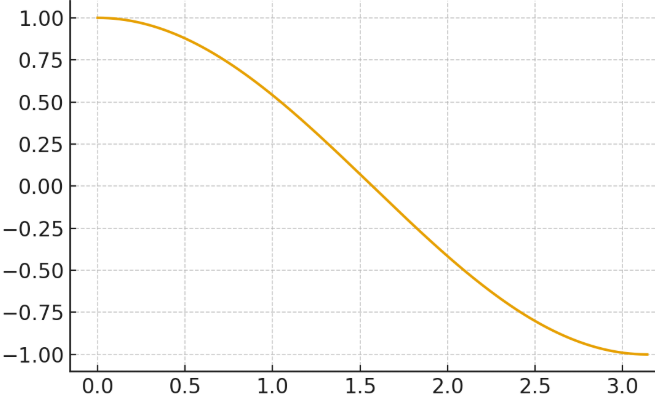}
        \caption{$\cos(\theta)$ on $[0,\pi]$.}
        \label{fig:cos_curve}
    \end{subfigure}
    \hfill
    \begin{subfigure}[b]{0.495\linewidth}
        \centering
        \includegraphics[width=\textwidth]{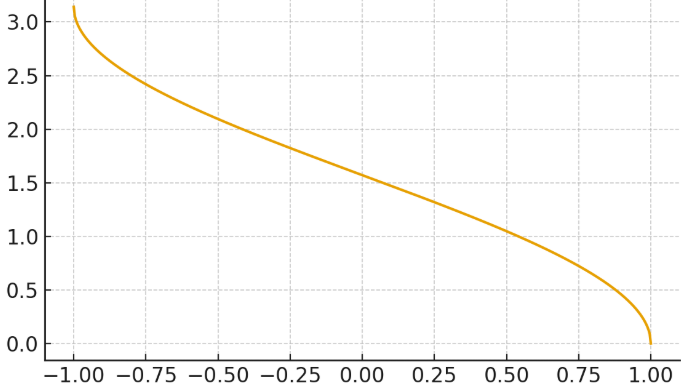}
        \caption{$\arccos(x)$ on $[-1,1]$.}
        \label{fig:arccos_curve}
    \end{subfigure}
    \caption{
    Illustration of the cosine function and its inverse. 
    }
    \label{fig:cos_arccos}
\end{figure}

\subsection{Extended details for MLR and IFA.}

\paragraph{MLR.} In the lifelong learning stage, we utilize the proposed Multimodal Latent Replay (MLR). Instead of storing raw sensory data, a compact buffer $\mathcal{B}$ maintains the multimodal latent representations ($\mathbf{H}$), which are concatenated from the frozen encoders, alongside the associated action ($a$) for training. In subsequent tasks, the stored $\mathbf{H}$ from previous tasks is fed into the temporal decoder for replay.

\paragraph{IFA.} To prevent interference (catastrophic forgetting) between the current task ($T_k$) and the previously learned tasks ($T_j$), we introduce the IFA. IFA's core mechanism encourages the global latent representation of the current task ($g_t(T_k)$) to remain closer to its own stable reference ($h^{(r)}(T_k)$) than to the references of previously learned tasks ($h^{(r)}(T_j)$).The IFA loss is explicitly defined to penalize violations of this distance constraint:

\begin{equation}
\label{eq:ara0}
\begin{aligned}
\mathcal{L}_{\text{IFA}} 
= \frac{1}{|\mathcal{P}|}
\sum_{\substack{(j,k)\in\mathcal{P} \\ j<k}}
\max \Big( 0,\;
& d\big(g_t(T_k), h^{(r)}(T_k)\big) \\
& {} - d\big(g_t(T_k), h^{(r)}(T_j)\big) + \delta 
\Big) \: , 
\end{aligned}
\end{equation}

The margin $\delta$ adapts to the relative positions of the task references themselves:

\begin{equation}
\label{eq:dist_delta0}
\begin{aligned}
\delta &= \alpha\, d\!\left(h^{(r)}(T_k),\, h^{(r)}(T_j)\right)
\end{aligned} \: ,
\end{equation}

\paragraph{Angle-based distance.}

The IFA loss is explicitly driven by the difference in distance between the global latent representation $g_t(T_k)$ and the task references: the positive reference $h^{(r)}(T_k)$ (for the current task $T_k$) and the negative references $h^{(r)}(T_j)$ (for the previous tasks $T_j$). Therefore, we compared how the standard cosine similarity and its angular counterpart (distance $d(\cdot, \cdot) = \arccos(\cdot)$) behave when used inside this crucial distance difference.

Fig.~\ref{fig:cos_arccos} illustrates this comparison.
The left panel plots $\cos(\theta)$ as a function of the angle $\theta \in [0,\pi]$ between two unit vectors, while the right panel shows $\arccos(x)$ as a function of $x \in [-1,1]$.
Cosine (left) quickly saturates in the high-similarity regime: when two representations are already very close, even noticeable angular changes produce almost no change in $\cos\theta$, making the resulting distance difference extremely small and hard to distinguish.
In contrast, the angular distance $\arccos(x)$ (right) expands this high-similarity region; small changes in $x$ near $1$ translate into clearly separable angular differences, providing more stable resolution across similarity levels and allowing IFA to better capture small deviations between task-level representations. Outside this range, $\arccos(x)$ behaves similarly to cosine similarity—differences between representations remain distinguishable.

\paragraph{Adaptive $\delta$.}

The choice of $\delta$ as a fixed hyperparameter would incur in a risk of choosing an overly large margin. Specifically, if $\delta$ is set larger than the distance between the two task references, i.e., $ \delta > d\big(h^{(r)}(T_k), h^{(r)}(T_j)\big)$, the IFA loss will always enforce a penalty.
This is guaranteed by the triangle inequality property of spherical geometry, which states that the difference between the distances to the two references is bounded by the distance between the references themselves:

\[
\begin{aligned}
d\big(g_t(T_k), h^{(r)}(T_k)\big)
&- d\big(g_t(T_k), h^{(r)}(T_j)\big) \\
&\le d\big(h^{(r)}(T_k), h^{(r)}(T_j)\big),
\end{aligned}
\]
If $\delta$ exceeds this maximum possible difference, the loss term inside the $\max(\cdot)$ will always be positive. This would enforce an unnecessary repulsive force, even when the current global latent representation $g_t(T_k)$ is already sufficiently closer to its own reference $h^{(r)}(T_k)$, potentially destabilizing learning.

To avoid this unnecessary penalty and allow the margin to adapt to the distance between references, we define $\delta$ dynamically:
\[
\delta = \alpha \, d\big(h^{(r)}(T_k), h^{(r)}(T_j)\big),
\quad \alpha \in (0,1).
\]

This guarantees that the margin is always bounded by the inter-reference distance, and scales proportionally to how far apart the tasks are in the angular space.

\section{Computational efficiency}
\label{SUPPMATsec:comp}

\begin{table}[t!]
\centering
\caption{Runtime and FLOPs comparison. See text for the explanation of Forward Time,
S-FLOPs, and T-FLOPs. All experiments are conducted on \textbf{LIBERO-GOAL}. $P$ means the probability of storing features in the buffer, which is ablated in Tab.~\ref{tab:ablation_buffer}. RR stands for raw replay, which stores past samples and replays them during training. Note that BASE stands for running our method without MLR or IFA. It is included solely to quantify the computational cost of the base architecture.}
\small
\resizebox{\linewidth}{!}{
\begin{tabular}{lcccc}
\toprule
\textbf{Method} 
& \textbf{$P$} 
& \textbf{Forward Time (ms)} 
& \textbf{S-FLOPs} 
& \textbf{T-FLOPs} \\
\midrule
BASE
& -
& 75.5 $\pm$ 2.43   &  $3.63\times 10^{11}$ & $4.36\times 10^{17}$ \\
BASE + RR  
& 0.5
& 75.8 $\pm$ 2.80 
& $3.63 \times 10^{11}$ & $12.96\times 10^{17}$\\
BASE + MLR 
& 0.5
& 75.5 $\pm$ 2.43 & $3.63\times 10^{11}$ & $7.62\times 10^{17}$ \\
BASE + MLR + IFA  
& 0.5
& 75.8 $\pm$ 2.80 
& $3.63 \times 10^{11}$ & $7.62\times 10^{17}$\\
\midrule
BASE + MLR  + IFA 
& 0.2
& 75.8 $\pm$ 2.43 
& $3.63 \times 10^{11}$ & $5.66\times 10^{17}$\\
BASE + MLR  + IFA 
& 0.1
& 75.8 $\pm$ 2.43 
& $3.63 \times 10^{11}$ & $5.01\times 10^{17}$\\
\bottomrule
\end{tabular}
}
\label{tab:Tab.SUPPMATsec:comp}
\end{table}

We report the computational efficiency using several primary metrics related to policy execution in Tab.~\ref{tab:Tab.SUPPMATsec:comp}. Forward Time refers to the time required to produce one action (i.e., a single forward pass). We computed forward time using a NVIDIA A100 GPU, by first running a warming up stage of 10 inferences (to avoid transient effects from initialization and caching), and then computing the average and standard deviation on 100 subsequent inferences.

The S-FLOPs correspond to the total number of Floating-Point Operations (FLOPs) required for a single forward pass of the policy to produce one action. This metric reflects model inference cost and is independent of data or training procedure. For completeness, we also report T-FLOPs, the total training FLOPs, computed as per-forward FLOPs multiplied by the total number of forward pass, backward pass, and gradient update during training (i.e., FLOPs × epochs × training sampless × 3). 

As shown in the result table, both MLR and IFA have a minimal overhead on
per-inference efficiency, indicating that their additional computations introduce
negligible runtime cost. In terms of total training FLOPs, adding MLR (P = 0.5) increases
the computational cost by roughly $0.75\times$ on \textbf{LIBERO-GOAL} due to replay, and the less the buffer is, the less the T-FLOPs. But this increase remains reasonable and acceptable given the resulting
performance gains compared to the performance in Tab.~\ref{tab:ablation_buffer}.

%% file: main.bib
@String(ICLR = {Int. Conf. Learn. Represent.})

@String(AAAI = {AAAI})

@String(ICLR  = {ICLR})

@article{loshchilov2017decoupled,
  title={Decoupled weight decay regularization},
  author={Loshchilov, Ilya and Hutter, Frank},
  journal={arXiv preprint arXiv:1711.05101},
  year={2017}
}

@article{liu2023libero,
  title={Libero: Benchmarking knowledge transfer for lifelong robot learning},
  author={Liu, Bo and Zhu, Yifeng and Gao, Chongkai and Feng, Yihao and Liu, Qiang and Zhu, Yuke and Stone, Peter},
  journal={Advances in Neural Information Processing Systems},
  volume={36},
  pages={44776--44791},
  year={2023}
}

@article{liu2023tail,
  title={Tail: Task-specific adapters for imitation learning with large pretrained models},
  author={Liu, Zuxin and Zhang, Jesse and Asadi, Kavosh and Liu, Yao and Zhao, Ding and Sabach, Shoham and Fakoor, Rasool},
  journal={arXiv preprint arXiv:2310.05905},
  year={2023}
}

@inproceedings{li2025robotic,
  title={Robotic visual instruction},
  author={Li, Yanbang and Gong, Ziyang and Li, Haoyang and Huang, Xiaoqi and Kang, Haolan and Bai, Guangping and Ma, Xianzheng},
  booktitle={Proceedings of the Computer Vision and Pattern Recognition Conference},
  pages={12155--12165},
  year={2025}
}

@inproceedings{yao2025think,
  title={Think Small, Act Big: Primitive Prompt Learning for Lifelong Robot Manipulation},
  author={Yao, Yuanqi and Liu, Siao and Song, Haoming and Qu, Delin and Chen, Qizhi and Ding, Yan and Zhao, Bin and Wang, Zhigang and Li, Xuelong and Wang, Dong},
  booktitle={Proceedings of the Computer Vision and Pattern Recognition Conference},
  pages={22573--22583},
  year={2025}
}

@article{zare2024survey,
  title={A survey of imitation learning: Algorithms, recent developments, and challenges},
  author={Zare, Maryam and Kebria, Parham M and Khosravi, Abbas and Nahavandi, Saeid},
  journal={IEEE Transactions on Cybernetics},
  year={2024},
  publisher={IEEE}
}

@article{zheng2022imitation,
  title={Imitation learning: Progress, taxonomies and challenges},
  author={Zheng, Boyuan and Verma, Sunny and Zhou, Jianlong and Tsang, Ivor W and Chen, Fang},
  journal={IEEE Transactions on Neural Networks and Learning Systems},
  volume={35},
  number={5},
  pages={6322--6337},
  year={2022},
  publisher={IEEE}
}

@article{tsuji2025survey,
  title={A Survey on Imitation Learning for Contact-Rich Tasks in Robotics},
  author={Tsuji, Toshiaki and Kato, Yasuhiro and Solak, Gokhan and Zhang, Heng and Petri{\v{c}}, Tadej and Nori, Francesco and Ajoudani, Arash},
  journal={arXiv preprint arXiv:2506.13498},
  year={2025}
}

@article{ding2023parameter,
  title={Parameter-efficient fine-tuning of large-scale pre-trained language models},
  author={Ding, Ning and Qin, Yujia and Yang, Guang and Wei, Fuchao and Yang, Zonghan and Su, Yusheng and Hu, Shengding and Chen, Yulin and Chan, Chi-Min and Chen, Weize and others},
  journal={Nature machine intelligence},
  volume={5},
  number={3},
  pages={220--235},
  year={2023},
  publisher={Nature Publishing Group UK London}
}

@article{xin2024parameter,
  title={Parameter-efficient fine-tuning for pre-trained vision models: A survey},
  author={Xin, Yi and Luo, Siqi and Zhou, Haodi and Du, Junlong and Liu, Xiaohong and Fan, Yue and Li, Qing and Du, Yuntao},
  journal={arXiv e-prints},
  pages={arXiv--2402},
  year={2024}
}

@article{mcinnes2018umap,
  title={Umap: Uniform manifold approximation and projection for dimension reduction},
  author={McInnes, Leland and Healy, John and Melville, James},
  journal={arXiv preprint arXiv:1802.03426},
  year={2018}
}

@article{lialin2023scaling,
  title={Scaling down to scale up: A guide to parameter-efficient fine-tuning},
  author={Lialin, Vladislav and Deshpande, Vijeta and Rumshisky, Anna},
  journal={arXiv preprint arXiv:2303.15647},
  year={2023}
}

@article{zheng2025towards,
  title={Towards lifelong learning of large language models: A survey},
  author={Zheng, Junhao and Qiu, Shengjie and Shi, Chengming and Ma, Qianli},
  journal={ACM Computing Surveys},
  volume={57},
  number={8},
  pages={1--35},
  year={2025},
  publisher={ACM New York, NY}
}

@inproceedings{wan2024lotus,
  title={Lotus: Continual imitation learning for robot manipulation through unsupervised skill discovery},
  author={Wan, Weikang and Zhu, Yifeng and Shah, Rutav and Zhu, Yuke},
  booktitle={2024 IEEE International Conference on Robotics and Automation (ICRA)},
  pages={537--544},
  year={2024},
  organization={IEEE}
}

@inproceedings{gao2021cril,
  title={CRIL: Continual robot imitation learning via generative and prediction model},
  author={Gao, Chongkai and Gao, Haichuan and Guo, Shangqi and Zhang, Tianren and Chen, Feng},
  booktitle={2021 IEEE/RSJ International Conference on Intelligent Robots and Systems (IROS)},
  pages={6747--5754},
  year={2021},
  organization={IEEE}
}

@article{lee2024incremental,
  title={Incremental learning of retrievable skills for efficient continual task adaptation},
  author={Lee, Daehee and Yoo, Minjong and Kim, Woo Kyung and Choi, Wonje and Woo, Honguk},
  journal={Advances in Neural Information Processing Systems},
  volume={37},
  pages={17286--17312},
  year={2024}
}

@inproceedings{roy2025m2distill,
  title={M2distill: Multi-modal distillation for lifelong imitation learning},
  author={Roy, Kaushik and Dissanayakc, Akila and Tidd, Brendan and Moghadam, Pcyman},
  booktitle={2025 IEEE International Conference on Robotics and Automation (ICRA)},
  pages={1429--1435},
  year={2025},
  organization={IEEE}
}

@article{rodriguez2018don,
  title={Don't forget, there is more than forgetting: new metrics for Continual Learning},
  author={Rodr{\'\i}guez, Natalia D{\'\i}az and Lomonaco, Vincenzo and Filliat, David and Maltoni, Davide},
  journal={CoRR},
  year={2018}
}

@article{zhu2022bottom,
  title={Bottom-up skill discovery from unsegmented demonstrations for long-horizon robot manipulation},
  author={Zhu, Yifeng and Stone, Peter and Zhu, Yuke},
  journal={IEEE Robotics and Automation Letters},
  volume={7},
  number={2},
  pages={4126--4133},
  year={2022},
  publisher={IEEE}
}

@article{chaudhry2019tiny,
  title={On tiny episodic memories in continual learning},
  author={Chaudhry, Arslan and Rohrbach, Marcus and Elhoseiny, Mohamed and Ajanthan, Thalaiyasingam and Dokania, Puneet K and Torr, Philip HS and Ranzato, Marc'Aurelio},
  journal={arXiv preprint arXiv:1902.10486},
  year={2019}
}

@article{lei2025dynamic,
  title={Dynamic Mixture of Progressive Parameter-Efficient Expert Library for Lifelong Robot Learning},
  author={Lei, Yuheng and Mao, Sitong and Zhou, Shunbo and Zhang, Hongyuan and Li, Xuelong and Luo, Ping},
  journal={arXiv preprint arXiv:2506.05985},
  year={2025}
}

@article{hu2022lora,
  title={Lora: Low-rank adaptation of large language models.},
  author={Hu, Edward J and Shen, Yelong and Wallis, Phillip and Allen-Zhu, Zeyuan and Li, Yuanzhi and Wang, Shean and Wang, Lu and Chen, Weizhu and others},
  journal={ICLR},
  volume={1},
  number={2},
  pages={3},
  year={2022}
}

@inproceedings{perez2018film,
  title={Film: Visual reasoning with a general conditioning layer},
  author={Perez, Ethan and Strub, Florian and De Vries, Harm and Dumoulin, Vincent and Courville, Aaron},
  booktitle={Proceedings of the AAAI conference on artificial intelligence},
  volume={32},
  number={1},
  year={2018}
}

@article{radford2019language,
  title={Language models are unsupervised multitask learners},
  author={Radford, Alec and Wu, Jeffrey and Child, Rewon and Luan, David and Amodei, Dario and Sutskever, Ilya and others},
  journal={OpenAI blog},
  volume={1},
  number={8},
  pages={9},
  year={2019}
}

@article{achiam2023gpt,
  title={Gpt-4 technical report},
  author={Achiam, Josh and Adler, Steven and Agarwal, Sandhini and Ahmad, Lama and Akkaya, Ilge and Aleman, Florencia Leoni and Almeida, Diogo and Altenschmidt, Janko and Altman, Sam and Anadkat, Shyamal and others},
  journal={arXiv preprint arXiv:2303.08774},
  year={2023}
}

@inproceedings{radford2021learning,
  title={Learning transferable visual models from natural language supervision},
  author={Radford, Alec and Kim, Jong Wook and Hallacy, Chris and Ramesh, Aditya and Goh, Gabriel and Agarwal, Sandhini and Sastry, Girish and Askell, Amanda and Mishkin, Pamela and Clark, Jack and others},
  booktitle={International conference on machine learning},
  pages={8748--8763},
  year={2021},
  organization={PmLR}
}

@article{hussein2017imitation,
  title={Imitation learning: A survey of learning methods},
  author={Hussein, Ahmed and Gaber, Mohamed Medhat and Elyan, Eyad and Jayne, Chrisina},
  journal={ACM Computing Surveys (CSUR)},
  volume={50},
  number={2},
  pages={1--35},
  year={2017},
  publisher={ACM New York, NY, USA}
}

@inproceedings{xie2024decomposing,
  title={Decomposing the generalization gap in imitation learning for visual robotic manipulation},
  author={Xie, Annie and Lee, Lisa and Xiao, Ted and Finn, Chelsea},
  booktitle={2024 IEEE International Conference on Robotics and Automation (ICRA)},
  pages={3153--3160},
  year={2024},
  organization={IEEE}
}

@inproceedings{jang2022bc,
  title={Bc-z: Zero-shot task generalization with robotic imitation learning},
  author={Jang, Eric and Irpan, Alex and Khansari, Mohi and Kappler, Daniel and Ebert, Frederik and Lynch, Corey and Levine, Sergey and Finn, Chelsea},
  booktitle={Conference on Robot Learning},
  pages={991--1002},
  year={2022},
  organization={PMLR}
}

@article{stepputtis2020language,
  title={Language-conditioned imitation learning for robot manipulation tasks},
  author={Stepputtis, Simon and Campbell, Joseph and Phielipp, Mariano and Lee, Stefan and Baral, Chitta and Ben Amor, Heni},
  journal={Advances in Neural Information Processing Systems},
  volume={33},
  pages={13139--13150},
  year={2020}
}

@inproceedings{kemker2018measuring,
  title={Measuring catastrophic forgetting in neural networks},
  author={Kemker, Ronald and McClure, Marc and Abitino, Angelina and Hayes, Tyler and Kanan, Christopher},
  booktitle={Proceedings of the AAAI conference on artificial intelligence},
  volume={32},
  number={1},
  year={2018}
}

@inproceedings{huang2024class,
  title={Class-incremental learning with clip: Adaptive representation adjustment and parameter fusion},
  author={Huang, Linlan and Cao, Xusheng and Lu, Haori and Liu, Xialei},
  booktitle={European Conference on Computer Vision},
  pages={214--231},
  year={2024},
  organization={Springer}
}

@article{pomerleau1988alvinn,
  title={Alvinn: An autonomous land vehicle in a neural network},
  author={Pomerleau, Dean A},
  journal={Advances in neural information processing systems},
  volume={1},
  year={1988}
}
